\documentclass[journal]{IEEEtran}
\usepackage{amsmath,amsfonts}
\usepackage{algorithmic}
\usepackage{algorithm}
\usepackage{array}
\usepackage[caption=false,font=normalsize,labelfont=sf,textfont=sf]{subfig}
\usepackage{textcomp}
\usepackage{stfloats}
\usepackage{url}
\usepackage{verbatim}
\usepackage{graphicx}
\usepackage{cite}
\usepackage{multirow}
\usepackage{float}
\usepackage{subcaption}
\usepackage[justification=centering]{caption}
\usepackage{tikz,xcolor}
\usepackage[
pdfauthor={derajan},
pdftitle={How to do this},
pdfstartview=XYZ,
bookmarks=true,
colorlinks=true,
linkcolor=blue,
urlcolor=blue,
citecolor=blue,
pdftex,
bookmarks=true,
linktocpage=true,   
hyperindex=true
]{hyperref}

\usepackage{amsmath}
\usepackage{hyperref}
\hypersetup{hypertex=true,
	colorlinks=true,
	linkcolor=blue,
	anchorcolor=blue,
	citecolor=blue}
\usepackage{indentfirst}
\usepackage{fontawesome}
\usepackage{nomencl}
\makenomenclature
\begin{document}

\title{Improving Question Embeddings with Cognitive Representation Optimization for Knowledge Tracing}

%

	\author{Lixiang Xu\href{https://orcid.org/0000-0001-8946-620X}{\includegraphics[scale=0.05]{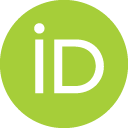}},
		Xianwei Ding\href{https://orcid.org/0009-0008-9931-8735}{\includegraphics[scale=0.05]{orcid.png}}, 
		Xin Yuan\href{https://orcid.org/0000-0001-5056-171X}{\includegraphics[scale=0.05]{orcid.png}},
		Zhanlong Wang\href{https://orcid.org/0009-0008-3210-6906}{\includegraphics[scale=0.05]{orcid.png}},
		Lu Bai\href{https://orcid.org/0000-0002-1033-8908}{\includegraphics[scale=0.05]{orcid.png}},
		Enhong Chen\href{https://orcid.org/0000-0002-4835-4102}{\includegraphics[scale=0.05]{orcid.png}},
		Philip S. Yu\href{https://orcid.org/0000-0002-3491-5968}{\includegraphics[scale=0.05]{orcid.png}},
		and Yuanyan Tang\href{https://orcid.org/0000-0002-6887-130X}{\includegraphics[scale=0.05]{orcid.png}}
	}

\maketitle

\begin{abstract}
	The Knowledge Tracing (KT) aims to track changes in students' knowledge status and predict their future answers based on their historical answer records. Current research on KT modeling focuses on predicting student' future  performance based on existing, unupdated records of student learning interactions. However, these approaches ignore the distractors (such as slipping and guessing) in the answering process and overlook that static cognitive representations are temporary and limited. Most of them assume that there are no distractors in the answering process and that the record representations fully represent the students' level of understanding and proficiency in knowledge. In this case, it may lead to many insynergy and incoordination issue in the original records. Therefore  we propose a Cognitive Representation Optimization for Knowledge Tracing (CRO-KT) model, which utilizes a dynamic programming algorithm to optimize structure of cognitive representations. This ensures that the structure matches the students' cognitive patterns in terms of the difficulty of the exercises. Furthermore, we use the co-optimization algorithm to optimize the cognitive representations of the sub-target exercises in terms of the overall situation of exercises  responses by considering all the exercises with co-relationships as a single goal. Meanwhile, the CRO-KT model fuses the learned relational embeddings from the bipartite graph with the optimized record representations in a weighted manner, enhancing the expression of students' cognition. Finally, experiments are conducted on three publicly available datasets respectively to validate the effectiveness of the proposed cognitive representation optimization model. The source code of CRO-KT is available at \url{https://github.com/bigdata-graph/CRO-KT}.
\end{abstract}



\printnomenclature

\nomenclature{$\alpha$}{It represents the difficulty difference between problems within the same knowledge point, usually set to be very large.}
\nomenclature{$\beta$}{It represents the difficulty difference between problems within the same knowledge point, usually set to be very small.}
\nomenclature{$J(S_1^k)$}{Cost function corresponding to the $k$-th state of the $i$-th exercise.}
\nomenclature{$J^*\big(S_1^k\big)$}{ Cost value of the $i$-th exercise at the $k$-th state has reached the minimum value within the specified range.}
\nomenclature{$E(x)$}{Objective function that includes set of all current sub-goals.} 
\nomenclature{$\gamma$}{Influence parameters between sub-goals.} 
\nomenclature{$S_{\varepsilon}$}{Student $\varepsilon$  is a student in the dataset sample.}
\nomenclature{$S_i^k$}{Represents the $k$-th state of the exercise in the $i$-th interaction of a student's interaction sequence.} 
\nomenclature{$u_i^k$}{Control value at the $k$-th state of the  $i$-th exercise.} 
\nomenclature{$\lambda_{ij}$}{The strength of cooperation that is  degree or intensity of collaboration or mutual support between individuals or entities.}  
\nomenclature{$\omega_{ij}$}{The strength of propagation  that is intensity or magnitude of the spread or transmission of something, such as influence, from one point.}


\nomenclature{$\mathcal{L}$}{The cross-entropy function between the generated relations of the problem space and skill space and the true relations.}

\section{Introduction}
\IEEEPARstart{K}{nowledge} Tracing (KT) \cite{19} uses students' historical question responses to predict their future performance. Earlier KT models, such as Bayesian Knowledge Tracking (BKT) \cite{17}, are based on the Hidden Markov Model. Although the BKT effectively captures changes in knowledge states during students' problem solving process, it does not account for differences in students' abilities and the discrepancy between different question containing the same knowledge points.\par
In 2015, Piech et al. \cite{18} firstly proposed the use of deep neural networks to construct knowledge tracking models, namely Deep Knowledge Tracing (DKT).  Zhang et al. \cite{29} proposed Dynamic Key Value Memory Network (DKVMN) model, which contains a static key matrix and a dynamic value matrix. Both DKT and DKVMN consider the hidden unit as the knowledge state of the student, but the performance of the above methods is not satisfactory when there is sparse historical data or when the student only interacts with a small number of knowledge points. \par
In order to solve the interaction problem of sparse data, the self-attentive mechanism has begun to receive attention. Pandey et al. \cite{4} proposed a KT framework based on self-attention mechanism, called Self-Attentive model for Knowledge Tracing (SAKT). This model constructs positional relationships between questions through self-attention mechanism, and uses it to enhance students' knowledge and the interaction and correlation between knowledge points. Along with that, other KT models based on attentional mechanisms have gradually emerged \cite{8}, \cite{15}, \cite{16}. Similarly Liu et al. \cite{11} proposed a pre-training model, namely Pre-training Embeddings via Bipartite Graph (PEBG), which constructs explicit and implicit relationships between questions and skills through a bipartite graph to train the embeddings.\par
As the field of KT continues to evolve, student personalisation factors and inter-student correlations are also considered in the models. For example, Long et al. \cite{12}  proposed a collaborative training model, called the Collaborative Embedding of Knowledge Tracing (CoKT), which combines inter-student and intra-student information to represent students' perspectives on a problem. Meanwhile, learning processes and behaviors have also received attention. With the development of machine translation, transformer-based \cite{23} self-attention models have been widely used, and KT models on Transformer \cite{9}, \cite{22}, \cite{25}, \cite{27}, \cite{30}  have proliferated in recent years. For example, Cui et al. \cite{5} proposed a transfomer-based model, called Multi-Relational Transfomer for Knowledge Tracking (MRT-KT), which reveals the complex cross-effects that exist between different question-response pairs in a sequence and enables the modeling of interactions between fine-grained question-response pairs. With the development of graph convolution and the application of heterogeneous graphs, convolution operations based on heterogeneity have also gained attention. For example, Wang et al. \cite{35} introduce a novel model called Feature Crosses Information-based KT (FCIKT) to explore the intricate interplay between questions, latent concepts, and question difﬁculties. FCIKT uses a fusion module to perform feature crossing on questions, integrating information from a multirelational heterogeneous graph with graph convolutional networks.\par
 \begin{figure*}[h]
	\centering
	\includegraphics[width=17cm,height=9cm]{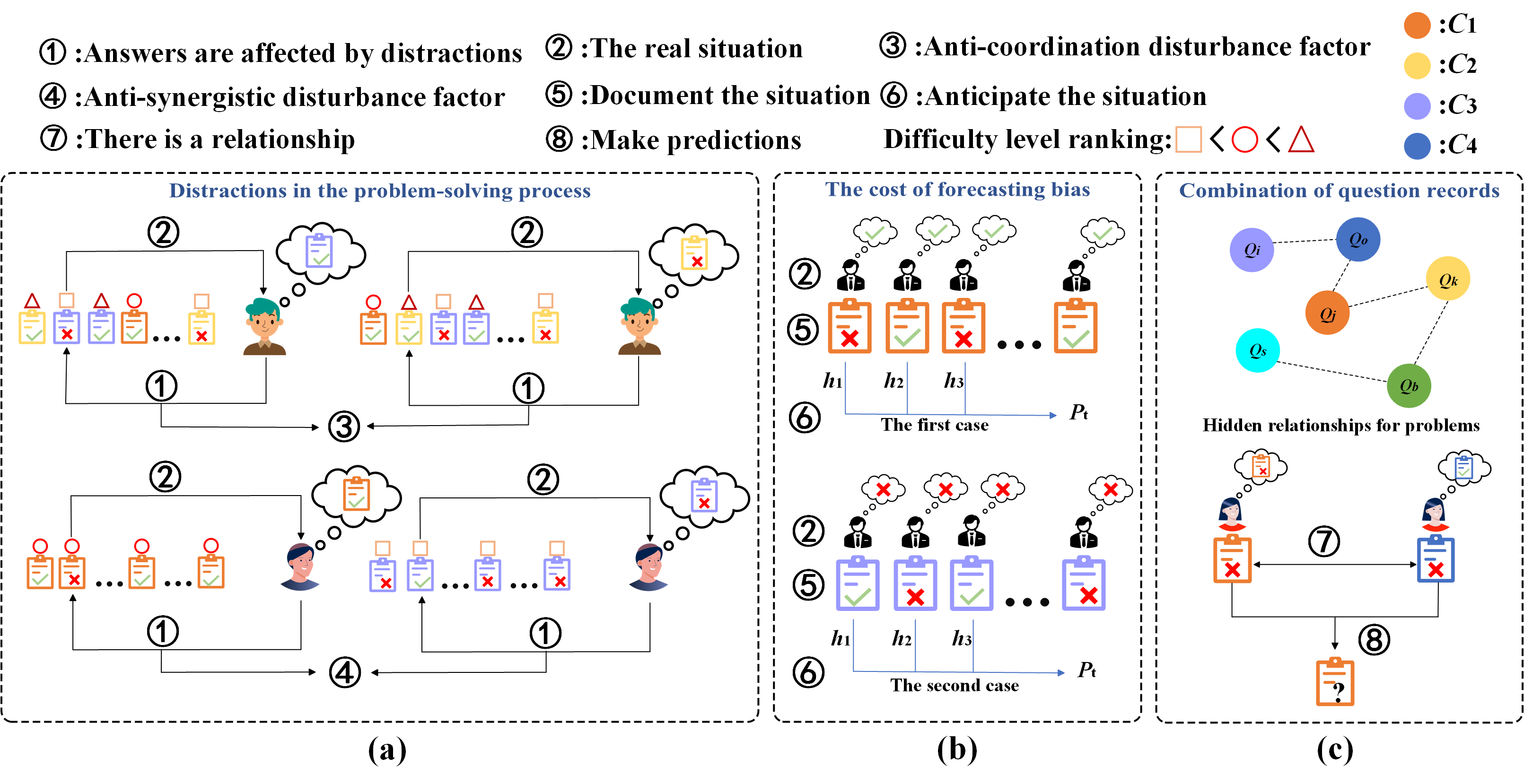}
	\caption{Taking the three parts of the figure from left to right as an example. Figure (a) shows the process of student working on the problem. Figure (b) shows the forecasting process of student performance. Figure (c) shows the process of predicting student performance based on the relationship between different questions.
	}
	\label{fig1}
\end{figure*}
However, most of the existing methods ignore impact of interfering factors such as slipping, guessing, problem solving habits, dependency tips and the structure of the questions themselves. Students are more susceptible to distractors during the initial stages of practice, and these distractors result in raw record representations that do not adequately represent student cognition. Therefore, cognitive representation needs to be optimized to closely approximate the actual cognition of students. In recent years, optimization methods have been widely applied in many fields, such as \cite{31}, \cite{32}, \cite{33}, \cite{34}; however, they have rarely been used in the field of KT. In this paper, we will apply optimization methods to KT to enhance the capabilities of KT models. A benign record representation should reflect the coordination and synergy within the record. Coordination means that the elements are appropriately paired with each other to achieve harmony, while synergy means that the elements are related and move in a common direction. However, there are many cases of uncoordination and lack of synergy in the original records. As shown in Fig. \ref{fig1}(a), there are two scenarios for problems with the same knowledge but different levels of difficulty: 1) If a harder exercise is written correctly afterwards, the student is more likely to have written the easier exercise correctly before, but the original record does not effectively characterize it. 2)  If an easier
exercise is written incorrectly afterwards, the student is more
likely to have written the harder exercise incorrectly before,
and the original record does not effectively characterize it. As shown in Fig. \ref{fig1}(a), for questions with the same knowledge and similar difficulty, there is a situation where the response status between questions should be  synergistic, but the recording representations are not effectively represented. Here, interference factors refer to elements that hinder a student's full expression. We use slipping  factors \cite{43}, \cite{44} and guessing factors \cite{45}, \cite{46} as examples to illustrate the issues of uncoordination and lack of collaboration. In the case of the first type of uncoordination, it may be caused by slipping factors, while the second type may result from guessing factors. At the same time, regarding the lack of collaboration between problems, we judge whether it is caused by  slipping factors or guessing factors based on the current problem, future related problems, and the answering situation of historical problems.  As shown  in Fig. \ref{fig1}(b), all previous models have relied on historical records to predict students' future performance, often over-exaggerating the representation of the original records regarding students' cognition. These models tend to ignore the impact of interfering factors and require significant resources and costs.\par
In the last few years, some models \cite{11}, \cite{35}  try to find associations between different questions and learn the relationships between them by some methods and eventually generate relational embeddings. However, they also overlook the interference students experience during the answering process, resulting in recorded representations of different questions that are not optimally combined. As shown  in Fig. \ref{fig1}(c), assuming there are problems involving concept $C_1$ and problems involving concept $C_4$, pairing problems with concept $C_4$ can aid in understanding problems with concept $C_1$. In terms of the record representations, for the problem with concept $C_1$, the record indicates that the student wrote the question incorrectly, and in fact the student was really unfamiliar with the question, and for the problem with concept $C_4$, the record indicates that the student wrote the question incorrectly, but in fact the student was familiar with and understood the question. So, even though we know the relationship between $C_1$ and $C_4$, we are still unclear about the student's actual mastery of $C_1$ and $C_4$. As a result, the predictions will still be bias.\par
In order to solve the above problems, we propose the CRO-KT model. To the best of our knowledge, we are the first to consider that primitive cognitive representations may be flawed, and introduce optimization \cite{10}, \cite{24} ideas and methods to address this problem. First, we designed the coordination module based on the dynamic programming algorithm. The coordination module of the CRO-KT model utilizes a dynamic programming algorithm to find the optimal solution of the problem representation based on students’ cognitive patterns and the overall distribution of their responses to questions with the same knowledge points but significant differences in difficulty levels. Secondly, we designed the collaboration module based on the co-optimization algorithm. There is a lack of synergy in the original records for these questions with the same knowledge and similar difficulty. Based on the correlation between these questions, the collaboration module of the CRO-KT model uses the co-optimization algorithm to iteratively optimize the question representation, ultimately finding the optimal solution to the problem representation. Then, the CRO-KT model also constructs relationships between questions through a bipartite graph, generates relational embeddings based on these relationships, and combines them with the optimized record representations so that the optimal solution representations of questions from different knowledge points can be combined together to express the students' cognitive. Finally, by conducting experiments on three publicly available datasets, we verified that the CRO-KT model effectively enhances students' cognitive representation. We also performed ablation experiments on the coordination module and the collaboration module, which further demonstrated the effectiveness and performance advancement in student cognitive representations for each module.

\section{Related Work}
\subsection{Traditional Methods and Deep Learning}
	KT methods can be categorized into traditional and deep learning based methods, in some cases traditional methods have comparable performance to deep learning based methods, but in other cases deep learning based methods are usually more powerful.\par
Most of the traditional methods are factor-based, they predict student reflections based on factors related to learning. The most classical method then is BKT \cite{17}, \cite{28}. It uses binary variables to represent students' knowledge states. BKT uses Hidden Markov Models to model knowledge states. Another typical type is the factor analysis approach \cite{20}, \cite{26}. The simplest model is Item Response Theory (IRT) \cite{6}. It measures the ability and difficulty of students to predict problems. Recent works have elaborated on the factors associated with learning. For example, Vie and Kashima \cite{20} introduced factors such as school ID, teacher ID, and they found that the performance of predicting student performance becomes better as the number of factors increases.\par
Most deep learning methods are state-based, and they use vectors to represent students' knowledge states. One representative method is DKT model, proposed by Piech et al. \cite{18}. DKT represents the student's knowledge state in terms of the hidden state of an LSTM and predicts the student's response by feeding the knowledge state to a binary classifier. As students interact with a problem, DKT updates their knowledge states through problem representations and student responses as their cognition improves. Many works have obtained better performance by extending DKT: Nagatani et al. \cite{14} considered forgetting behaviour; Chen et al. \cite{3} labelled prerequisite relationships between concepts; Su et al. \cite{21} encoded problem embeddings with textual descriptions, and Liu et al. \cite{11} pre-trained problem embeddings. Unlike DKT-based methods that use a single vector to represent a student's knowledge state, there are methods that use multiple vectors to represent the knowledge state of different concepts. One such approach is the DKVMN \cite{29}. The DKVMN stores concept representations in a key matrix and knowledge states in a value matrix. In addition, many works follow DKVMN. For instance, Abdelrahman and Wang \cite{1} incorporate LSTM and memory networks in knowledge state updating.\par
\subsection{Learning Gains and Learning Behaviors}
The learning processes and behaviors of students have also received attention. For example, Shen et al. \cite{36,39} introduced Learning Process Consistent Knowledge Tracing (LPKT), a method that directly models students' learning processes to monitor their knowledge states. Specifically, it formalizes basic learning units as tuples of ``practice-answer-time-answer.'' The model deeply measures learning gains and their variations based on the differences between the current learning unit and previous ones, the interval time, and the relevant knowledge states of the students. However, this model does not account for potential interference that students may face. For example, students may exhibit slipping or guessing when answering, which could lead to a situation where the positive gain described by LPKT is actually a negative gain, or where the negative gain described is actually a positive gain. Meanwhile, Xu et al. \cite{40} proposed Learning Behavior-oriented Knowledge Tracing (LBKT), which explicitly investigates the impact of learning behaviors on students' knowledge states. This model first analyzes and summarizes several key learning behaviors, including speed, attempts, and hints during the learning process. Given the significant differences in the characteristics of these behaviors, LBKT quantitatively assesses their effects on knowledge acquisition. Additionally, considering the close relationships among different learning behaviors, LBKT captures complex dependency patterns to evaluate their combined effects and comprehensively updates students' knowledge acquisition while accounting for forgetting factors. Although LBKT considers the influence of learning behaviors on cognitive states to some extent, the behaviors it describes remain static and temporary, failing to sufficiently consider the potential interference that students might encounter. This shortcoming may result in an incomplete reflection of students' actual behaviors, leading to information gaps. \par
\subsection{Knowledge Structure and Knowledge Crossing}
With the development of KT and graph neural networks, the structure of knowledge concepts and information propagation has received widespread attention. For instance, Tong et al. \cite{38} proposed a new framework called Structured Knowledge Tracing (SKT), which utilizes various relationships within the knowledge structure to simulate the influence propagation between concepts. In the SKT framework, it not only considers the temporal effects of exercise sequences but also the spatial effects of the knowledge structure. It employs two novel formulas to simulate the influence propagation on a knowledge structure with multiple relationships. For undirected relationships, such as similarity, a synchronous propagation method is used, allowing influence to propagate bidirectionally between adjacent concepts. Wang et al. \cite{35} introduced a new model called Feature-Crossing Information-based Knowledge Tracing (FCIKT) to explore the intricate interactions among questions, latent concepts, and question difficulty. FCIKT uses a fusion module to perform feature-crossing operations on questions and leverages graph convolutional networks to integrate information from the multi-relational heterogeneous graph we constructed. It deploys a multi-head attention mechanism to enrich the static embedding representations of questions and concepts with dynamic semantic information, simulating real-world problem-solving scenarios.\par

\subsection{Cognitive Representation Optimization}
The CRO-KT model consists of two main modules: the coordination and collaboration modules. The coordination module uses dynamic programming algorithms to find optimal solutions, analyzing problem difficulty and answering status to resolve inconsistencies. For questions with the same knowledge point, if there is a large difficulty difference and discrepancies in answers, the coordination module adjusts problems dynamically to ensure effective coordination. The collaboration module relies on collaborative optimization algorithms to maintain consistency in answers for related problems, optimizing the system's sub-goals. This optimization ensures that problems with similar difficulty and the same knowledge points receive more consistent answers.
\subsection{Summary}
Although previous methods have achieved good results, they assumed that students were not affected by interfering factors in the process of solving problems, and that the records fully represented their level of understanding and proficiency, overstating the representativeness of the original records. In this paper, we will use the CRO-KT model to effectively reduce which record representations are inaccurate according to the overall situation of students' answers using an optimisation algorithm, and use the bipartite graph to learn the implicit and explicit relationships between questions and skills, and then combine them with the optimal solutions of the record representations to perform the optimal solution combination to express more effectively students' cognitive situation. 

\section{ PRELIMINARY}
	\subsection{ Sequence Generation}
According to research and experiments in the field of educational psychology \cite{2}, \cite{13}, \cite{7}, the level of students' cognition is positively correlated with difficulty. The more difficult the exercise, the greater the ability to overcome easier exercise and the higher the cognitive level. At the same time, this reflects the factors and potential traits related to students' problem-solving abilities, such as IQ, knowledge base, and learning background. The familiarity and understanding of knowledge points by students depend more on their individual traits and abilities, so students' cognitive states do not necessarily correlate positively with time. The quality of questions answered by students (the difficulty of the questions themselves) is positively correlated with students' cognitive states and can effectively distinguish between students.  This indicates that students have the capability and potential to overcome easier exercises. Our difficulty calculation process is as follows:
\begin{equation}
	D_i=\frac1{P_i}~,P_i=\frac{C_i}{M_i}~,i=1,2,3,\ldots,
\end{equation}
where $D_i$ represents the difficulty corresponding to question $i$ and $P_i$ represents the percentage of correct answers corresponding to question $i$. Question $i$ may be interacted least once by at least one student. $M_i$ specifically refers to the total number of times all students answered question $i$, while $C_i$ represents the number of correct answers out of the total number of times question $i$ was answered. The ratio of $C_i$ to $M_i$ is $P_i$, and the inverse of $P_i$ is $D_i$, where the larger the $P_i$ implies the higher the rate of correctness, the simpler the question is, and at the same time, the smaller its inverse is, implying the less difficult the question is. On the contrary, if the smaller $P_i$ means the lower the correct rate and the more difficult the question is, then its inverse is also larger, which means the more difficult it is. We need to collect the set of students $S$ and the set of questions $Q$. Taking student $\varepsilon$ as an example, according to the mapping relationship between questions and students, we get the set $S_{\varepsilon}$=$\{(q_{{a}},r_{\varepsilon}^{1}, K_{{q_{a}}}),(q_{{b}},r_{\varepsilon}^{2},K_{{q_{b}}}),(q_{{c}},r_{\varepsilon}^{3},K_{{q_{c}}}),\ldots\}.$ Our model needs to correspond to the knowledge of the question as well as the answer situation. In addition, we need the difficulty of the question as well. So the final set is $S_{\varepsilon}$=$\{(q_{a},r_{\varepsilon}^{1},K_{q_{a}},D_{q_{a}}),(q_{b},r_{\varepsilon}^{2},K_{q_{b}},D_{q_{b}}),(q_{c},r_{\varepsilon}^{3}, K_{q_{c}},D_{q_{c}}),\ldots\}$. It is simplified as  $S_{\varepsilon}$=$\{(e_{\varepsilon}^{1},r_{\varepsilon}^{1},k_{\varepsilon}^{1},d_{\varepsilon}^{1}),(e_{\varepsilon}^{2},r_{\varepsilon}^{2},k_{\varepsilon}^{2},d_{\varepsilon}^{2}),\ldots, (e_{\varepsilon}^{t},r_{\varepsilon}^{t},k_{\varepsilon}^{t},d_{\varepsilon}^{t})\}$. Take student $\varepsilon$ as an example, if the knowledge point  in the $i$-th interaction is $k_\varepsilon^i$, and  in the $u$-th interaction is $k_\varepsilon^u$, where $k_\varepsilon^i=k_\varepsilon^u$, the difficulty of their questions are $d_{\varepsilon}^i$ and $d_{\varepsilon}^u$, with $i<u$. If there is a significant difference between $d_{\varepsilon}^i$ and $d_{\varepsilon}^u$ , then, depending on the responses $r_\varepsilon^i$ and $r_\varepsilon^u$ two uncoordinated cases arise: i) $r_\varepsilon^i=0,r_\varepsilon^u=1,d_\varepsilon^i<d_\varepsilon^u$ and $|d_\varepsilon^i-d_\varepsilon^u|\geq\alpha_.$ ii) $r_\varepsilon^i=1$, $r_{\varepsilon}^{u}=0,d_{\varepsilon}^{i}>d_{\varepsilon}^{u}$ and $|d_{\varepsilon}^{i}-d_{\varepsilon}^{u}|\geq\alpha,$ where $\alpha$ represents a significant difference in question difficulty. If the difficulties $d_\varepsilon^i$ and $d_\varepsilon^u$ are similar but the answers $r_\varepsilon^i$ and $r_\varepsilon^u$ are different, this reflects a lack of synergy in the record. In response to these uncoordinated and insynergy elements in the original records, we will elaborate on the optimization algorithm modules addressing these issues in section \ref{section4}.

\begin{figure*}[h]
	\centering
	\includegraphics[width=14.5cm,height=7.7cm]{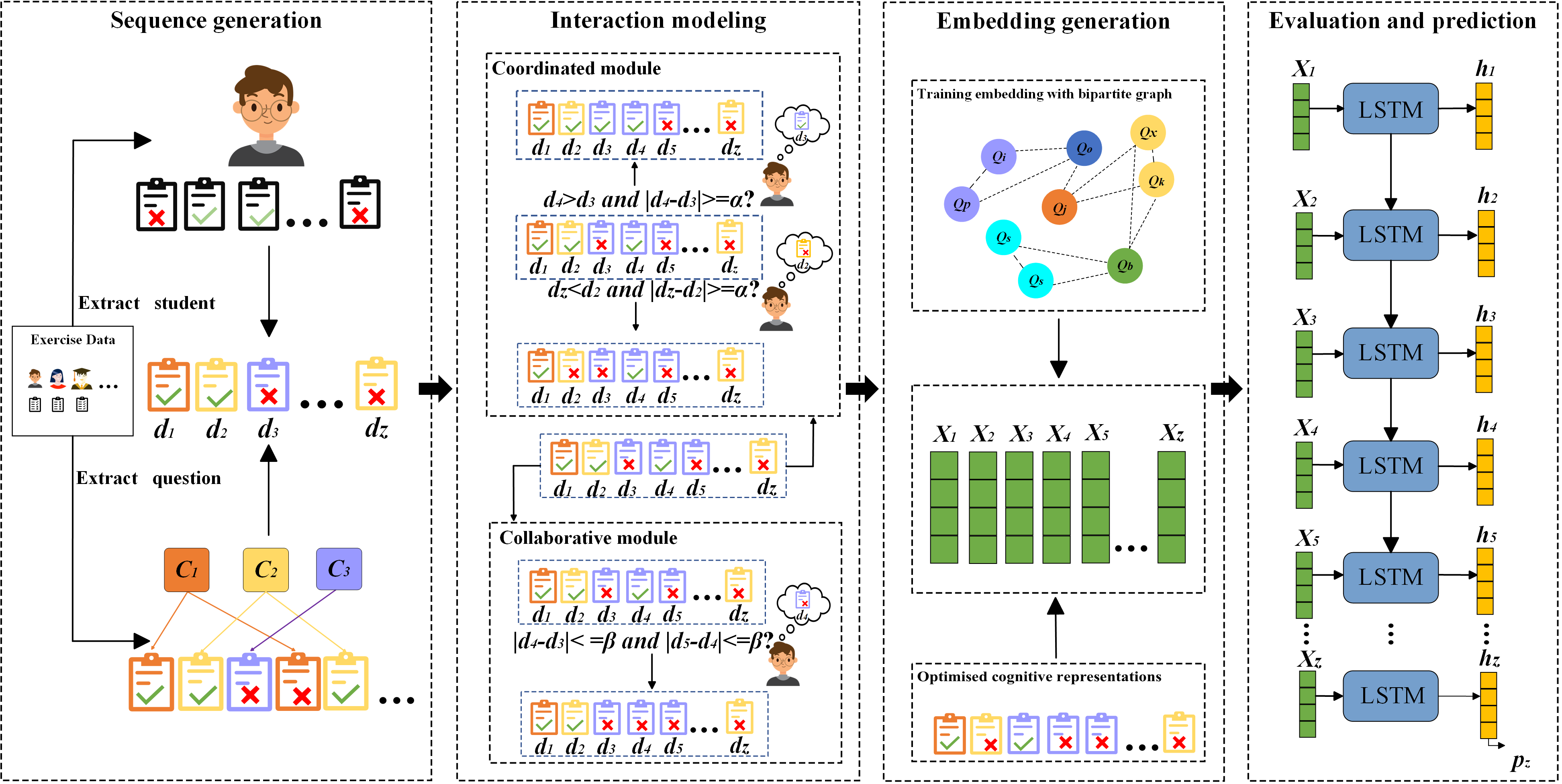}
	\caption{Schematic framework diagram of the CRO-KT model.}
	\label{fig2}
\end{figure*}
\subsection{Training Embedding with Bipartite Graph}
To establish the relationships between problems with different
knowledge points, we learn these relationships through a bipartite
graph.
\begin{equation}
	\hat{r}_{ij}=\sigma({q}_i^T{s}_j),i=1,\ldots,M,j=1,\ldots,N .
\end{equation}
First we generate the relation $\hat{r}_{ij}$ based on the learnable problem space ${q}_i$ and skill space ${s}_j$, and then apply the sigmoid function to obtain probabilities. We then construct the cross entropy function based on the problem and skill relations $\hat{r}_{ij}$ in the bipartite graph, and train $q_i$ and $s_j$ to approximate the relationships in the bipartite graph by minimizing the cross entropy,
\begin{equation}
	\mathcal{L}_1({Q},{S})=\sum_i^M\sum_j^N-\left(r_{ij}log\hat{r}_{ij}+(1-r_{ij})log(1-\hat{r}_{ij})\right).
\end{equation}
The next step will be to train the relationships between questions and problems as well as between skills and skills, in the same way,
\begin{equation}
	\hat{r}_{ij}^q=\sigma({q}_i^T{q}_j),i,j\in[1,\ldots,M],
\end{equation}
\begin{equation}
	\hat{r}_{ij}^s=\sigma(s_i^Ts_j),i,j\in[1,\ldots,N],
\end{equation}
where $q$ and $s$ are the problem space and skill space, respectively. Then the cross entropy function is constructed to learn the true relationship $r_{ij}^q$ and $r_{ij}^s$,
\begin{equation}
	\mathcal{L}_2({Q})=\sum_i^M\sum_j^M-\left(r_{ij}^qlog\hat{r}_{ij}^q+(1-r_{ij}^q)log(1-\hat{r}_{ij}^q)\right),
\end{equation}
\begin{equation}
	\mathcal{L}_3({S})=\sum_i^N\sum_j^N-\left(r_{ij}^slog\hat{r}_{ij}^s+(1-r_{ij}^s)log(1-\hat{r}_{ij}^s)\right).
\end{equation}
It is easy to see through the bipartite graph that $r_{ij}^q=1$ if there is an intersection of the knowledge points for problem ${q}_i$ and problem ${q}_j$ and 0 otherwise. Similarly $r_{ij}^\mathrm{s}=1$ if there is an intersection of the corresponding problems for skills
${s}_i$ and ${s}_j$ and 0 otherwise. In addition to this, we also add the attribute features of the problems themselves for training,
\begin{equation}
	\mathcal{L}_4({Q},{S},{\theta})=\sum_{i=1}^{|Q|}(a_i-\hat{a}_i)^2,
\end{equation}
where $\hat{a}_i$ are trainable attribute features and $a_i$ are attribute features of the actual problem. Finally, we train by joint optimization,
\begin{equation}
	\min_{{Q},{S},{\theta}}\lambda\big(\mathcal{L}_1({Q},{S})+\mathcal{L}_2({Q})+\mathcal{L}_3({S})\big)+(1-\lambda)\mathcal{L}_4({Q},{S},{\theta}).
\end{equation}
The final embedding is generated, where $\lambda$ is the trade-off coefficient. The final embedding is weighted and fused with the optimized record representations as input for prediction and evaluation.

\section{THE CRO-KT METHOD}
\label{section4}
The CRO-KT model is divided into two main modules, the coordination module and the collaboration module. The coordination module is mainly based on the dynamic programming algorithm to find the optimal solution of each question record that meets the coordination relationship. Our framework diagram is shown in Fig. \ref{fig2}.  The collaborative module is mainly based on the collaborative optimization algorithm, which makes the answers to the questions with approximate relationships consistent with each other, so as to achieve the optimal solution of the sub-objective.
\subsection{Coordination Module}
\begin{figure*}[h]
	\centering
	\includegraphics[width=7.9cm,height=8.7cm]{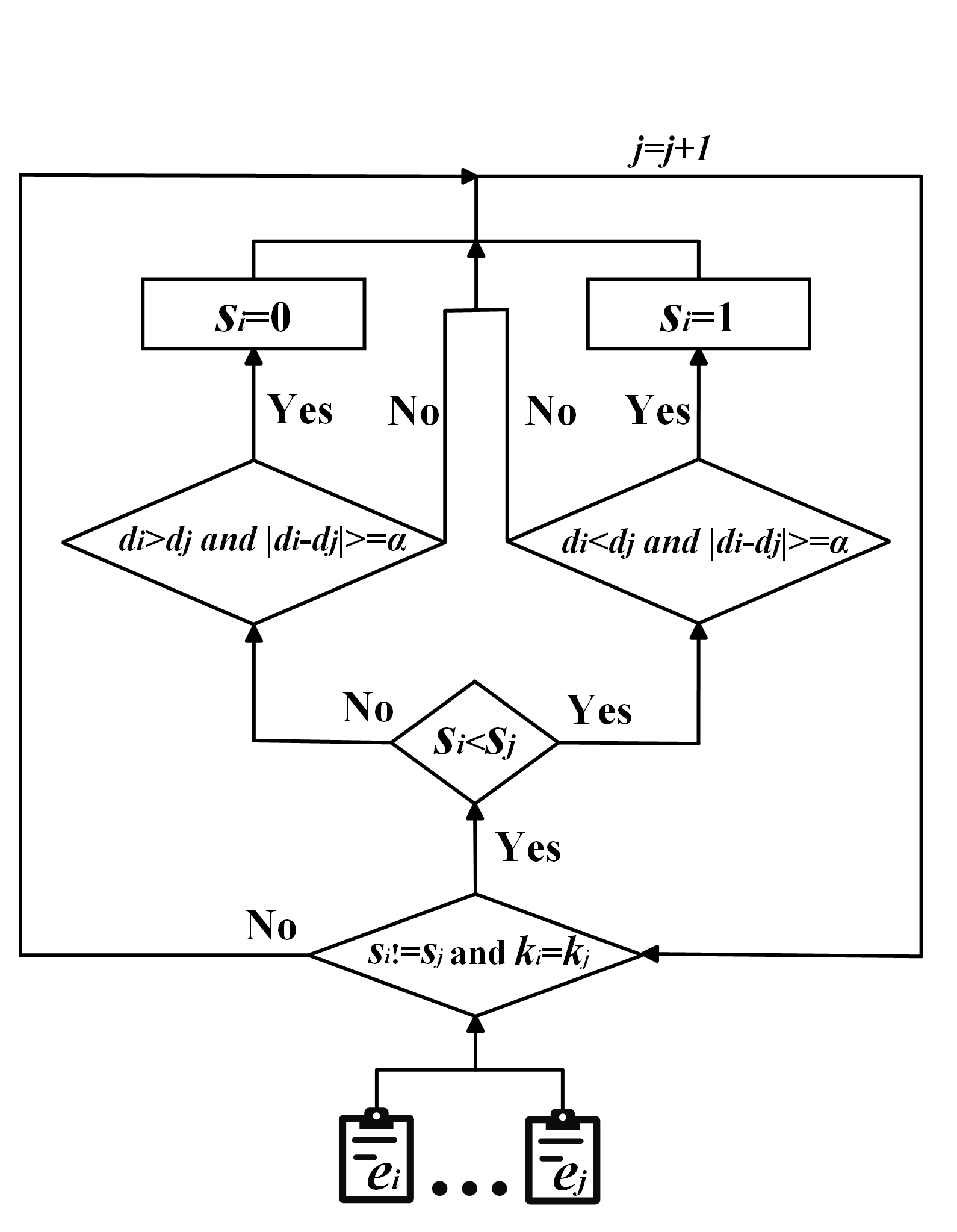}
	\caption{Algorithm flowchart.}
    \label{fig3}
\end{figure*}

The coordination module focuses on the discovering incoherence between questions by comparing their difficulty and the question answering status. Incoherent records occur in sequences with the same knowledge points where the absolute value of their difficulties difference is greater than or equal to $\alpha$ and the question response status $S$ is different. Our algorithm flowchart is shown in Fig. \ref{fig3}.    Taking student $\varepsilon$ as an example,  $S_{\varepsilon}$= $\{(e_\varepsilon^1,r_\varepsilon^1,k_\varepsilon^1,d_\varepsilon^1),(e_\varepsilon^2,r_\varepsilon^2,k_\varepsilon^2,d_\varepsilon^2),\ldots,(e_\varepsilon^t,r_\varepsilon^t,k_\varepsilon^t,d_\varepsilon^t)\}$. We take out the sequences with the same knowledge points from the sequence $S_\varepsilon$ and keep only their answer states (0 or 1) and the difficulty of the corresponding exercises, and finally get the state-difficulty sequence $I$= $\{(S_1^0,\mathrm{df}_1),(S_2^0,\mathrm{df}_2),\ldots,(S_i^0,\mathrm{df}_i),\ldots,(S_n^0,\mathrm{df}_n)\}.$ 
Considering discrete sequence nonlinear systems,
\begin{equation}
	S_i^{k+1}=F\big(S_i^k,u_i^k\big),k=0,1,2,\ldots,
	\label{eq10}
\end{equation}
where $k$ describes the steps of the system trajectory and $S_i^k\in$ $\{0,1\}$ is the state variable of the system, which represents the $k$-th state of the $i$-th exercise. The $u_i^k$ function as follows:
\begin{equation}
	u_i^k=v\big(S_i^k,S_{k+i+1}^0,\mathrm{df}_i,\mathrm{df}_{k+i+1}\big),k=0,1,2,\ldots~.
\end{equation}
Let $i$=1, then $S_1^{k+1}$=$F(S_1^k,u_1^k),u_1^k$=$v$ $(S_1^k,S_{k+2}^0,\mathrm{df}_1,\mathrm{df}_{k+2})$, where $u_1^k\in\{0,1\}$ is the control variable of the system, representing the control value at the $k$-th state of the first exercise. We need the answer state, difficulty of the first exercise, and the initial answer state and difficulty of the later exercises to output the values of the control variables via the function $\upsilon$:
\begin{equation}
	\small
	\hspace{-3mm}
	\left.v=\left\{\begin{array}{c}1,S_1^k<S_{k+2}^0 ~and~ \mathrm{df}_1<\mathrm{df}_{k+2} ~and~ |\mathrm{df}_1-\mathrm{df}_{k+2}|\geq\alpha\\-1,S_1^k>S_{k+2}^0 ~and~ \mathrm{df}_1>\mathrm{df}_{k+2} ~and~ |\mathrm{df}_1-\mathrm{df}_{k+2}|\geq\alpha\\0\end{array}\right.\right.
	\!.
\end{equation}
Eq. (\ref{eq10}) describes a discrete  nonlinear system in the general sense of a nonlinear system with an affine form, namely
\begin{equation}
	S_1^{k+1}=f\big(S_1^k\big)+g\big(S_1^k\big)u_1^k,
\end{equation}
where $g(S_1^k)$ is the control function and $f(S_1^k)=S_1^k, g(S_1^k)=1$, which ultimately gives $S_1^{k+1}=$ $S_1^k+u_1^k,$ and $u_1^k$ is the amount of control to change the state of $S_1^k$. If $u_1^k=1$, it indicates the first case of incoherent record representation. Obviously $S_1^k$ $<S_{k+2}^0$, and $S$$\in\{0,1\}$, so $S_1^k=0,~S_{k+2}^0=1$, and its next state is $S_1^{k+1}=0+1=1.$ If $u_1^k=-1$, it indicates that the second case of incoherent record representation  occurs. Obviously $S_1^k>S_{k+2}^0$, and $S$$\in \{ 0, 1\}$, so $S_1^k= 1,~S_{k+ 2}^0= 0$, and its next state is $S_1^{k+1}=$1-1=0. If $u_1^k=0$, the record representation coordination, and its next state remain unchanged as $S_1^{k+1}=S_1^k+0\overset{}{\operatorname*{=}}S_1^k.$ The corresponding cost function is as follows:
\begin{equation}
	J(S_1^k)=\sum_{n=k}^N\gamma^{p-k}U(S_1^p,u_1^p).
	\label{eq14}
\end{equation}
We minimize the cost of lack of coordination due to flaws in the record by changing the response state of the question. Let $U$ be the effective function and $\gamma$ be the discount factor, and let $0<\gamma\leq1$. The valid functions are as follows:
\begin{equation}
	U(S_1^p,u_1^p)=|(S_1^p-u_1^p)-S_1^p|=|-u_1^p|.
\end{equation}
Starting from the $k$-th state of the first exercise, we calculate the number of record incoherencies that occur further along. A higher number indicates a more inaccurate record representation and a greater cost, while a lower number represents a more accurate record and a lesser cost, with the optimal cost being zero. Starting from the $k$-th state of the first exercise, count the number of times that the recording representation appears incoherent going forward. More frequent occurrences are more costly, while fewer occurrences are less costly. According to the optimal control theory of dynamic programming, the optimal cost function as follows:
\begin{equation}
	J^*\big(S_1^k\big)=\min_{u_1^k,u_1^{k+1},\ldots,u_1^N}\sum_{p=k}^N\gamma^{p-k}U\big(S_1^p,u_1^p\big).
\end{equation}
It can be written as:
\begin{equation}
	\begin{aligned}
		&J^{*}\big(S_{1}^{k}\big)=\min_{u_{1}^{k}}\{U\big(S_{1}^{k},u_{1}^{k}\big)+\gamma\min_{u_{1}^{k+1},u_{1}^{k+2},\ldots,u_{1}^{N}}\sum_{p=k+1}^{N}\\
		&\gamma^{p-k-1}U\big(S_{1}^{p},u_{1}^{p}\big)\}.
	\end{aligned}
\end{equation}

Thus, $J^*(S_1^k)$ satisfies the discrete sequence Hamilton-Jacobi-Bellman (HJB) equation:
\begin{equation}
	J^*\big(S_1^k\big)=\min_{u_1^k}\{U\big(S_1^k,u_1^k\big)+\gamma J^*\big(S_1^{k+1}\big)\}.
\end{equation}
The corresponding optimal control \cite{24} is as follows:
\begin{equation}
	u^*\big(S_1^k\big)=\arg\min_{u_1^k}\{U\big(S_1^k,u_1^k\big)+\gamma J^*\big(S_1^{k+1}\big)\}.
\end{equation}
The corresponding optimal state is as follows:
\begin{equation}
	S^*\big(S_1^k\big)=\arg\min_{S_1^k}\{U\big(S_1^k,u_1^k\big)+\gamma J^*\big(S_1^{k+1}\big)\}.
\end{equation}
When the optimal state is obtained, the optimal value of the cost function is 0. From Eq. (\ref{eq14}), the expansion can be carried out:
\begin{equation}
	\small
	J\big(S_1^k\big)=U\big(S_1^k,u_1^k\big)+\gamma^1U\big(S_1^{k+1},u_1^{k+1}\big)+\gamma^2U\big(S_1^{k+2},u_1^{k+2}\big)+\dots~.
\end{equation}
Assuming $S_{1}^{\theta }$ is optimal, then $u_{1}^{\theta }= 0, u_{1}^{\theta + 1}= 0,\ldots,u_{1}^{n}= 0.$ If $u_{1}^{\theta }= 0$, then $U( S_{1}^{\theta }, u_{1}^{\theta }) =  | ( S_{1}^{\theta }- u_{1}^{\theta } ) - S_{1}^{\theta } |=  | S_{1}^{\theta }- S_{1}^{\theta } | = 0.$ At this point $S_{1}^{\theta + 1}= S_{1}^{\theta }$, the state is unchanged. $u_1^{\theta+1}=0,\mathrm{~then~}U\left(S_1^{\theta+1},u_1^{\theta+1}\right)=0,S_1^{\theta+2}=S_1^{\theta+1}$. Finally,  $J\left(S_1^k\right)=0+0+0+\ldots=0$, representing that the function has reached a minimum. At this point $S_1^\mathrm{\theta}$ is the optimal state and $u_{1}^\mathrm{\theta}$ is the optimal control. Therefore, $S_1^\theta$ is the optimal solution of the first exercise state we want.
\subsection{Collaboration Module }
\begin{figure*}[h]
	\centering
	\includegraphics[width=7.9cm,height=7.5cm]{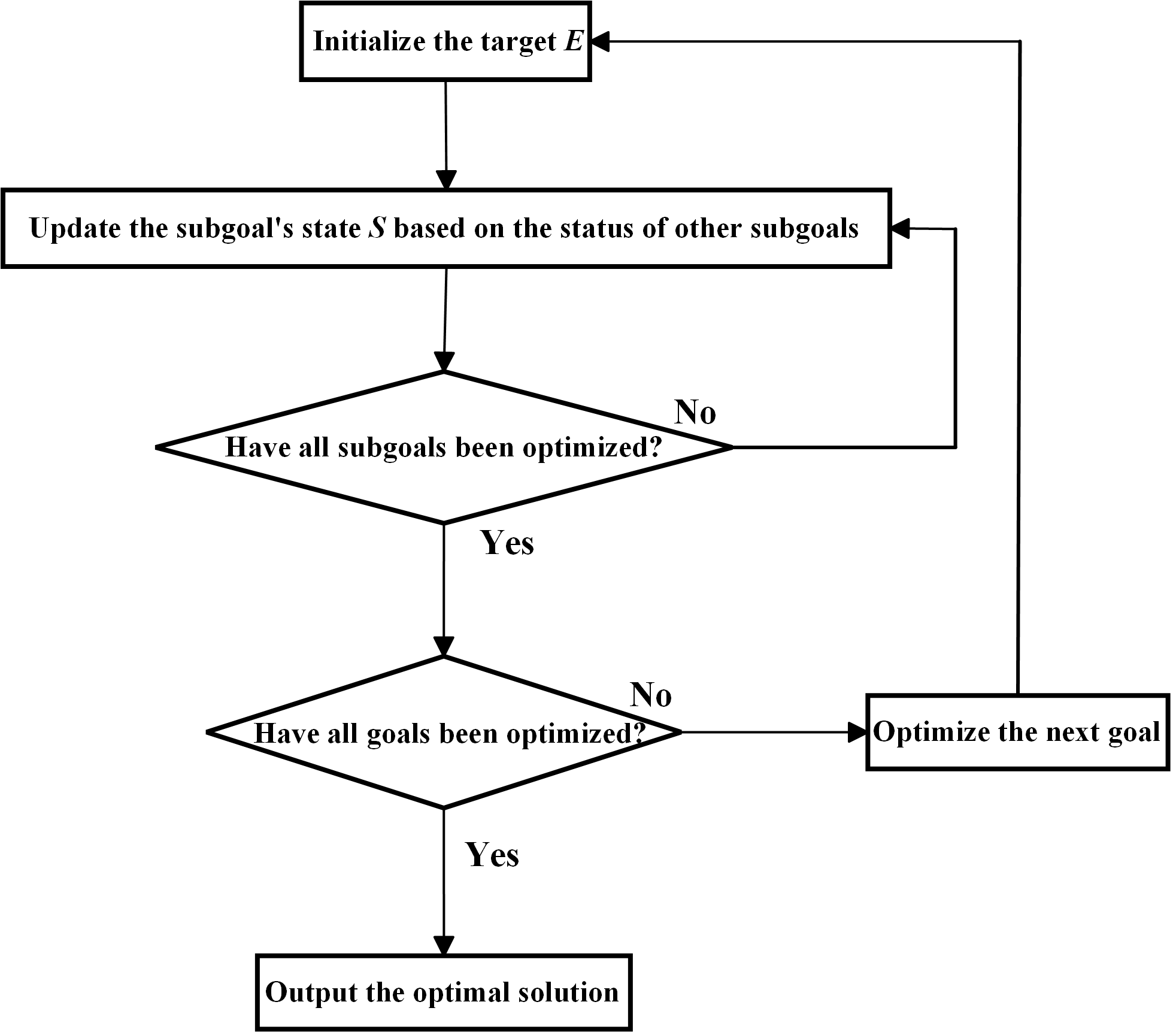}
	\caption{Algorithm flowchart.}
	\label{fig16}
\end{figure*}

Co-optimization module is  based on the synergistic relationship between problems. In general, an objective or a set can be divided into related subclasses or sub-objectives that share similarity, correlation, and consistency. Specifically, let the objective function be $E(x)$.  The collaborative optimization algorithm decomposes it into $m$ simple sub-objective functions, ensuring that the absolute value of the difficulty difference for each sub-objective exercise is less than  $\beta$. $\beta$ is a very small number, which represents the similarity in difficulty between the problems. Our algorithm flowchart is shown in Fig. \ref{fig16}  
\begin{equation}
	E(x)=E_1(x^1)+E_2(x^2)+...+E_i(x^i)+...+E_m(x^m),
\end{equation}
where $x^1 \sim x^m$ have the same knowledge and are exercises of similar difficulty.
\begin{equation}
	E_i^0(x^i)=d_{i}S_i^0,
\end{equation}
where $d_{i}$ is the difficulty of the $i$-th exercise, $S_i^0$ is the initial state of the answer of the $i$-th exercise. Typically, only the subfunction of the current objective function is optimized once. We set the coefficient $\gamma$, where $\gamma$ is the coefficient of discriminative consistency. Specifically, $\gamma=\gamma_1/\gamma_2$ , with 
$\gamma_1$ being the numerator of discriminative consistency, also known as the effective factor, which represents the direction of consistency of the sub-target answer state. The more subgoal exercises answered correctly, the greater $\gamma_1$ and the greater the likelihood of a cognitive representation of 1. The more subgoal exercises answered incorrectly, the smaller $\gamma_1$ becomes and the greater the likelihood of a cognitive representation of 0. We design $\gamma_1$ according to the subgoals, as follows:
\begin{equation}
	\gamma_1=(1-\lambda_{ij})E_i^0\big(x^i\big)+\lambda_{ij}\sum_{j\neq i}\omega_{ij}E_j^0\big(x^j\big),
	\label{eq24}
\end{equation}
where $\lambda_{ij}$ and $\omega_{ij}$ are weighting coefficients satisfying $0\leq\lambda_{ij}\leq1,\omega_{ij}\leq1$. Here, $\lambda_{ij}$ represents the strength of cooperation (set as $\lambda_{ij} = 1/2$), and $\omega_{ij}$ portrays the strength of propagation, with $\omega_{ij}=1/L_{ij}$, where $L_{ij}$ is the propagation distance between the two exercises. The larger the distance, the smaller $\omega_{ij}$, indicating less propagation strength; conversely, the smaller the distance, the larger 
$\omega_{ij}$, indicating greater propagation strength. However, there is a difference in propagation strength depending on the exercise before $i$ and after $i$. If $j$ is before $i$, the state of exercise $j$ cannot be directly verified by the state of exercise $i$ because of the presence of temporal affective factors, such as forgetting, so $L_{ij}=|i-j|.$ If $j$ comes after $i$, the state of exercise $j$ can directly verify the state of exercise $i$, so $L_{ij}=1.$ Eventually  Eq. (\ref{eq24}) simplifies to
\begin{equation}
	\gamma_1=\frac12d_{i}S_i^0+\frac12\sum_{j\neq i}\frac1{L_{ij}}d_{j}S_j^0,L_{ij}=\left\{\begin{matrix}1,i<j\\|i-j|,i>j\end{matrix}\right..
\end{equation}
In order to discriminate and quantify the direction of the sub-goal states, we designed
the overall influence factor $\gamma_2$, as the denominator,
\begin{equation}
	\gamma_2=\frac12d_{i}+\frac12\sum_{j\neq i}\frac1{L_{ij}}d_{j},L_{ij}=\left\{\begin{matrix}1,i<j\\|i-j|,i>j\end{matrix}\right..
\end{equation}
Finally we calculate the discriminant coefficient $\gamma=\gamma_1/\gamma_2$, and we get the
optimization results for the sub-objectives in the current objective,
\begin{equation}
	E_i^1\big(x^i\big)=d_{i}S_i^1,S_i^1=\big\{\begin{matrix}1,\gamma\geq0.5\\0,\gamma<0.5\end{matrix}\big.~~.
\end{equation}
After all the sub-objectives in the current objective are optimized, then the next objective is optimized. Before optimization, it is necessary to set the sub-objective optimization results of the previous objective to the initial state, i.e. $S_i^0=S_i^1,~E_i^0(x^i)=d_{{i}}S_i^0$, and then the next objective function is as follows:
\begin{equation}
	\small
	E(x)=E_1(x^1)+E_2(x^2)+...+E_i(x^i)+...+E_m(x^m)+E_{m+1}(x^{m+1}).
\end{equation}
The same optimization is carried out again until the last problem with the same knowledge points and similar difficulty, the final optimization result is the optimal solution, and the optimization results of all sub-objectives are related to the optimization results of other sub-objectives.

\subsection{Evaluation and Prediction}
The training completed through the bipartite graph constructs relationships between different skill-related problems, ultimately combining the optimised cognitive records to form an optimal combination of solutions to characterizes the student's cognition. Specifically our assessment task is to do a weighted fusion of the optimised cognitive representations with the training embeddings from the bipartite graph. The hidden layer $h_t$ obtained by taking $x_{t}$ as input is used to generate the predicted result $\hat{y}_{t+1}$, which is then compared with the true result $y_{t+1}$, the more similar the two are, the better the performance is.

\section{Experiment}
In this section, we conduct experiments to evaluate the performance of our proposed model, as well as the performance of the ablation experiments of each module.
\subsection{Dataset}
We use three real-world datasets, and the statistics for the three datasets are shown in Table \ref{table1}.
\begin{table}[h]
	\caption{Statistics of the three datasets.}
	\centering     
	\begin{tabular}{cccl}
		\hline
		&ASSIST09&ASSIST12&EdNet\\
		\hline
		students & 3,841& 27,405&5,000\\
		questions& 15,911&	47,104&	11,775\\
		skills & 123&	265&	1,837\\
		records & 190,320&	1,867,167&	1,156,254\\
		\hline
		\label{table1}
	\end{tabular}
\end{table}
Both ASSIST09 and ASSIST12 were collected from the ASSISTments online tutoring platform.  For both datasets, we delete records in the absence of skill and scaffolding issues. We also remove users with less than three records. 

\begin{itemize}
	\item {\textbf{ASSIST09\footnote{Available:\url{https://sites.google.com/site/assistmentsdata/home/assistment-2009-2010-data}}}}: It contains 123 skills, 15, 911 questions answered by 3, 841 students, and a total of 190, 320 records.
	\item {\textbf{ASSIST12\footnote{Available:\url{https://sites.google.com/site/assistmentsdata/home/2012-13-school-data-with-affect}}}}: It contains 265 skills, 47, 104 questions answered by 27, 405 students, and 1, 867, 167 records.
	\item {\textbf{EdNet\footnote{Available:\url{https://github.com/riiid/ednet}}}}: It contains 1837 skills, 11, 775 questions answered by 5, 000 students, 1, 156, 254 records.
\end{itemize}
\par
In the ASSIST09 and ASSIST12 datasets, most students tend to perform well on certain knowledge points (i.e., they have mastered these points), while a smaller group of students fail to master them. This distribution leads to an imbalance in the data, with the ``mastered'' category significantly outweighing the ``not mastered" category. In this case, the model is prone to bias towards predicting students as having mastered the knowledge points, neglecting those who have not.
\par
In the EdNet dataset, there is a significant variation in how students master different knowledge points. Some knowledge points are mastered by the majority of students, while others are only mastered by a few. As a result, the ``mastered" category for certain knowledge points may dominate the majority of the samples, while categories like ``not mastered" have fewer samples. This data imbalance can lead to the model over-predicting students' mastery of knowledge points, neglecting those who have not yet mastered them.

\subsection{Baselines}
We present all our baselines as follows:\\
\begin{itemize}
	\item {\textbf{DKT}}  \cite{18} uses recurrent neural networks to model student skill learning.
	\item {\textbf{DKVMN}} \cite{29}  uses a key-value memory network to store the underlying conceptual representation and state of skills.
	\item {\textbf{SAKT}} \cite{4}  proposed a self-attention based approach to determine the correlation between KCs.
	\item {\textbf{PEBG+DKT}} \cite{11}  introduced bipartite graph's to establish explicit and implicit relationships between problems and skills.
	\item {\textbf{CoKT}} \cite{12}  combines inter-student and intra-student information to represent students' perceptions of problems.
	\item {\textbf{MRT-KT}} \cite{5}  reveals complex cross-effects between response states corresponding to different questions in a sequence, and enable interaction modelling between fine-grained question response states.
	\item {\textbf{FCIKT}} \cite{35}  better integrates question information through feature crossing and graph convolution networks, using multi-head attention mechanisms and GRU to dynamically update students' knowledge states for effective knowledge tracing.
\end{itemize}

\subsection{Experimental Setup}
We discuss the excellence of our model in terms of AUC \cite{41}  and ACC \cite{42} , based on the previously shown knowledge tracking model in Table \ref{table2}. CRO-KT has only a few hyperparameters. The dimension of vertex features dv is set to 64. The final question embeddings dimension d = 128. We use the Adam algorithm to optimize our model, and mini-batch size for three datasets is set to 256, the learning rate is 0.001. We also use dropout with a probability of 0.5 to alleviate overfitting. We divide each dataset into 80$\%$ for training and validation, and 20$\%$for testing.

	\begin{table*}[h]
		
		\caption{Comparison of all the models.}
		\vspace{-0.6em}                     
		\centering 
		\setlength{\tabcolsep}{12pt}                                     
		\begin{tabular}{cccccccccc}                         
			\hline                                 
			\multirow{2}{*}{Model}&                      
			\multicolumn{3}{c}{\textbf{ASSIST09}}&            
			\multicolumn{3}{c}{\textbf{ASSIST12}}&
			\multicolumn{3}{c}{\textbf{EdNet}}\cr     
			&AUC &ACC &RMSE&AUC &ACC &RMSE&AUC &ACC &RMSE\\
			\hline                                      
			DKT&0.7356&  0.7179& 0.4366&	0.7013&  0.6842& 0.4226&		0.6909&  0.6889& 0.4700\\
			DKVMN&0.7394&  0.7076& 0.4416&	0.6752&  0.7048& 0.4224&	0.6893&  0.6660& 0.4538\\
			SAKT&0.7894&  0.7649& 0.4270&	0.7206&  0.7306& 0.4236&	0.6929&  0.6879& 0.4781\\
			PEBG+DKT&0.8287&  0.7669& 0.4200&	0.7665&  0.7423& 0.415&	0.7765&  0.7076& 0.4600\\
			CoKT&0.7682&  0.7324& 0.4296&	0.7401&  0.7380& 0.4206&	0.7374&  0.6887& 0.4584\\
			MRT-KT&0.8223&  0.7841& 0.4180&	0.7698&  0.7544& 0.4100&	0.7753&  \textbf{0.7319}& 0.4550\\
			FCIKT&0.7912&  0.7500& 0.4250&	0.8020&  0.7650& 0.4080&	0.7463&  0.7200& 0.4650\\
			\hline
			CRO-KT&\textbf{0.8782}&  \textbf{0.8183}& \textbf{0.4011}&  \textbf	{0.8051}&   \textbf{0.7812}&  \textbf{0.4002}& \textbf{0.7880}	&  0.7258& \textbf{0.4500}\\
			\hline
			\label{table2}
		\end{tabular}
	\end{table*}

\begin{figure*}[h]
	\centering
	\includegraphics[width=17cm,height=6.6cm]{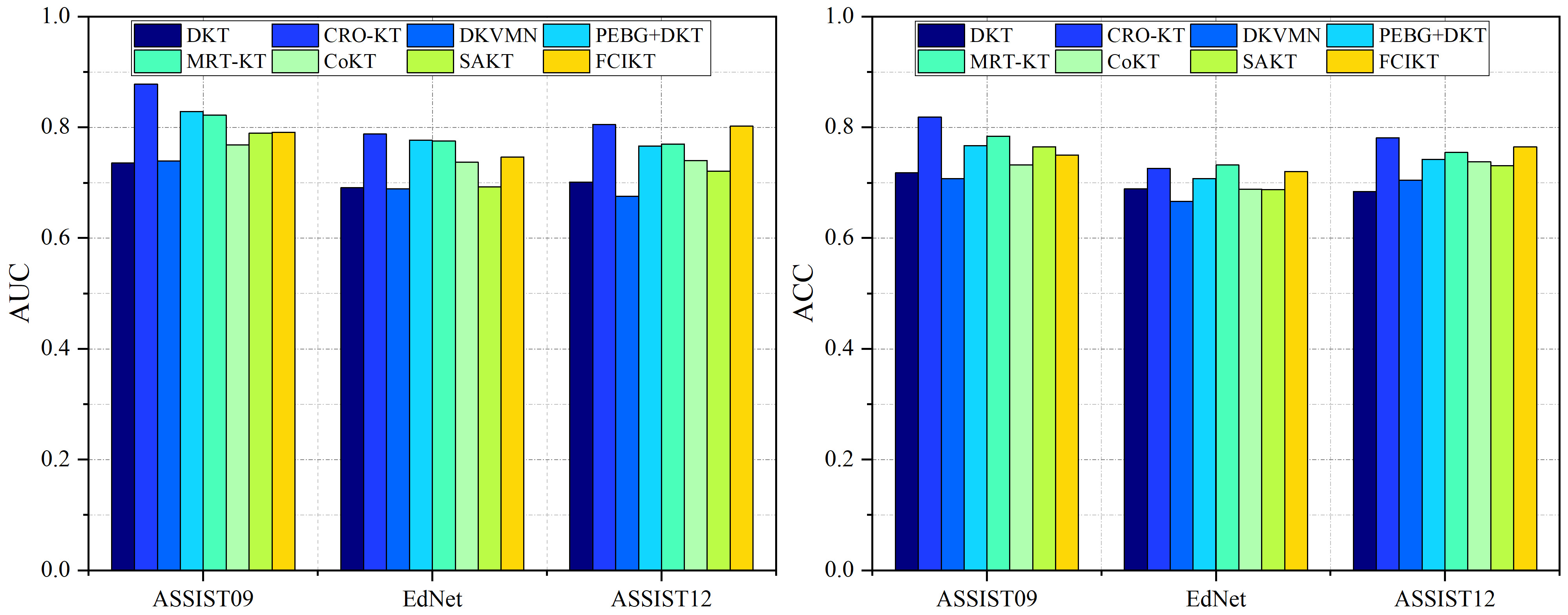}
	\caption{Bar graphs show the performance of each model.}
	\label{fig4}
\end{figure*}

As shown in Fig. \ref{fig4}, the performance of our model is significantly better than the performance of the other models. Compared to the EdNet dataset, ASSIST09 has a smaller number of questions and requires much less knowledge. This often leads to interactions involving questions of the same knowledge, thereby increasing the potential for performance improvement. Compared to the ASSIST09 dataset, EdNet has a shorter sequence of interactions per student and a very large number of knowledge points. This results in fewer interactions involving questions of same knowledge, but it provides a richer optimal composition of cognitive representation, which also increases the potential for performance improvement.\par
\begin{figure*}[h]
	\centering
	\includegraphics[width=17cm,height=6.0cm]{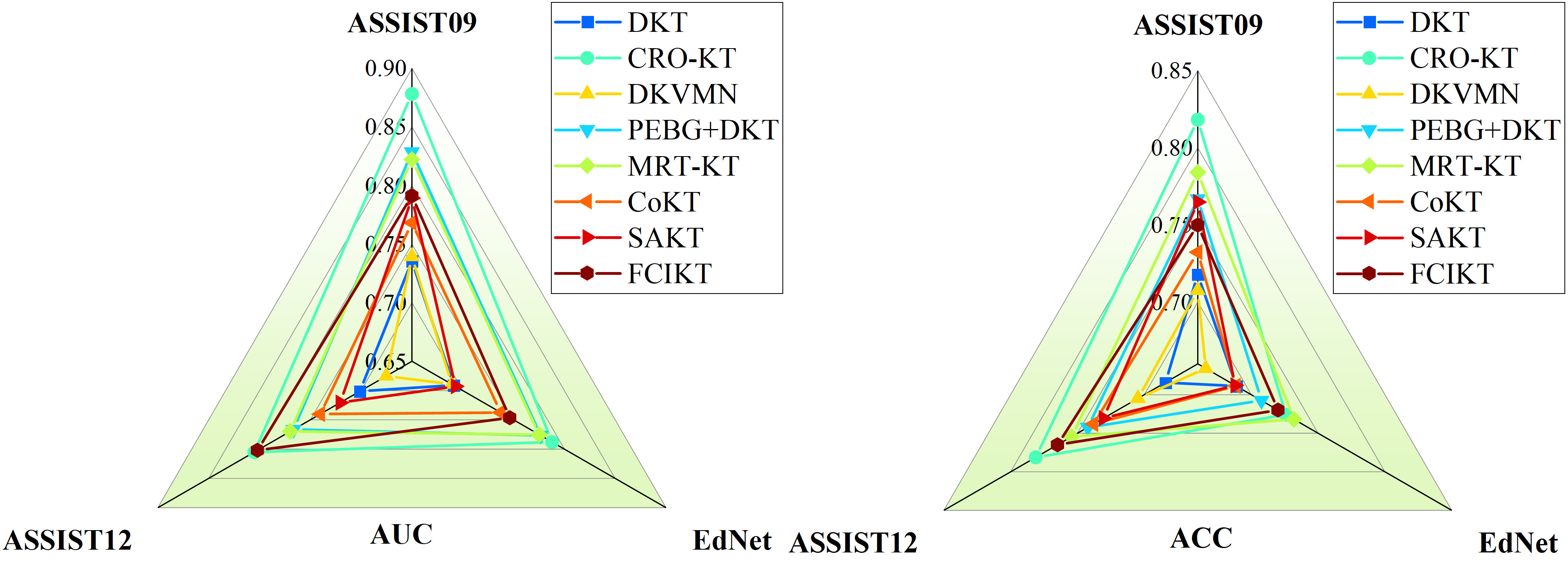}
	\caption{Radar chart showing performance of each model.}
    \label{fig5}
\end{figure*}

Similar to ASSIST09, the ASSIST12 dataset has a sparse number of knowledge points. However, ASSIST12 features a very large number of questions and interactions. Since the same knowledge point information is often encountered during interactions, it also increases the possibility of improving performance. As shown in Fig. \ref{fig5}, the distribution of the radar charts shows that even though the characteristics of the dataset itself lead to different accuracy improvements, the overall situation is still in a relatively smooth state.\par
\begin{figure*}[h]
	\centering
	\includegraphics[width=17cm,height=6.9cm]{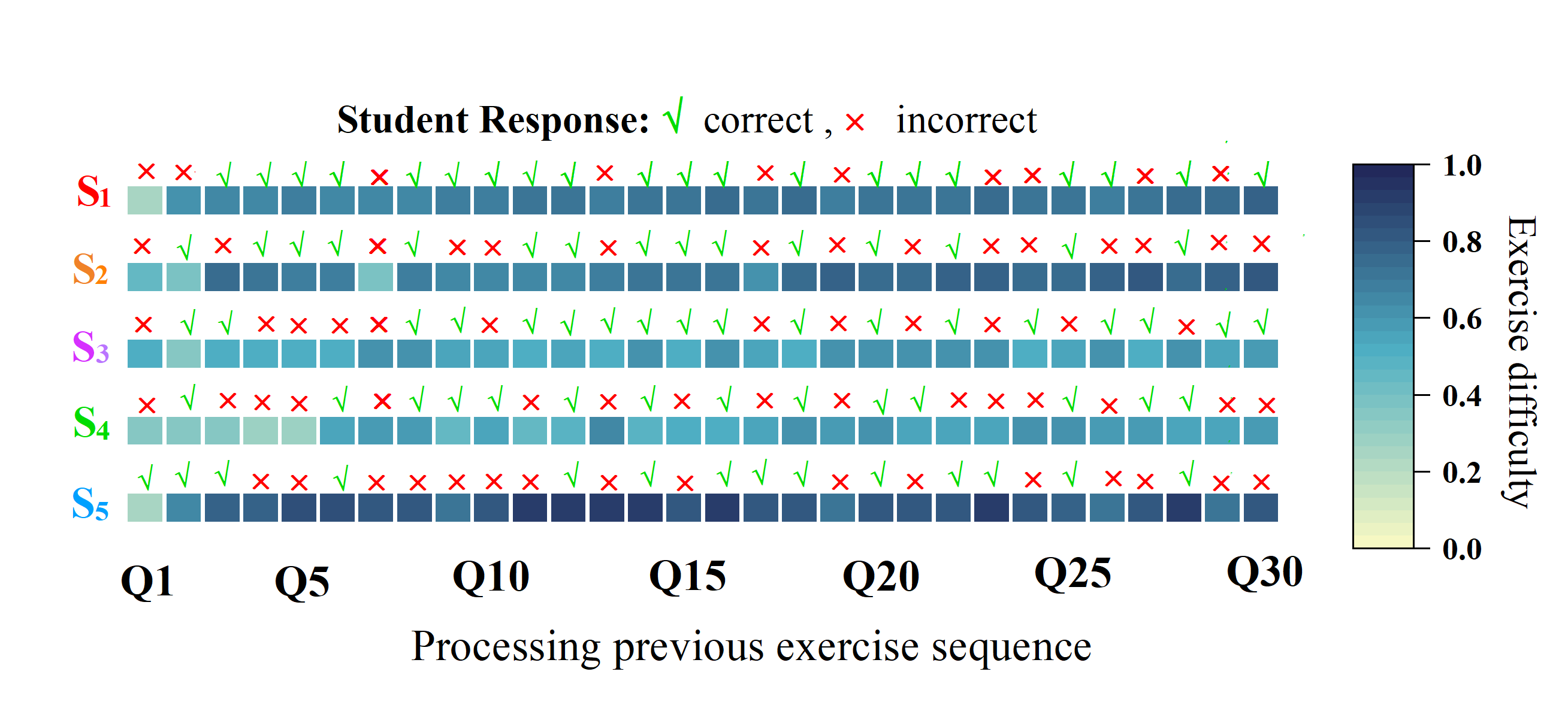}
	\caption{ Processing  exercise before.}
	\label{fig6}
\end{figure*}

\begin{figure*}[h]
	\centering
	\includegraphics[width=17cm,height=6.9cm]{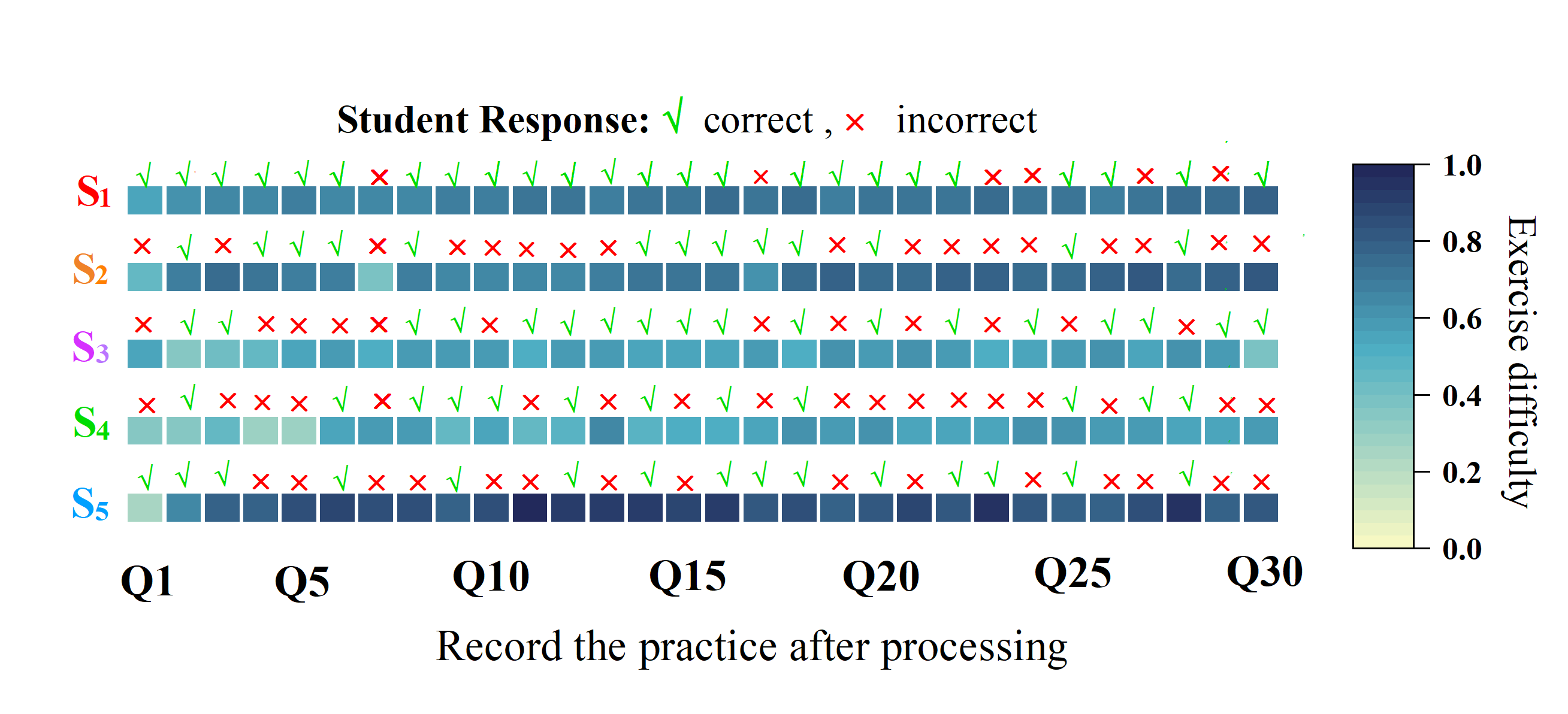}
	\caption{Processing exercise after.}
	\label{fig7}
\end{figure*}

In addition to this, we designed an experiment before and after the treatment of students' exercise records to highlight our innovations; we selected five students $S_1$ to $S_5$ and they had a high number of interactions. We selected their performance in each of the 30 exercises in their own interactive exercises, which all contain the same knowledge points and vary in difficulty. As shown in Fig. \ref{fig6}, it is the record of the exercises before we processed them. Students interacted with the exercises independently of each other and there was variability in their performance. We set a colour representation for each exercise and normalized the difficulty. Darker colours represent higher difficulty and lighter colours represent lower difficulty. After processing through our model, the results are shown in Fig. \ref{fig7}. 
\begin{table*}[h]
	\caption{A comparison table of the answering situation before and after.}
	\centering   
	\label{table20}
	\setlength{\tabcolsep}{2pt}  
	\label{tab:freq}
	\begin{tabular}{ccccccccccc}
		\hline
		\multicolumn{11}{c}{Pre-optimization of cognitive representation}\\
		\hline
		$S_1$&$Q_1$(\faTimes,0.15)&$Q_2$(\faTimes,0.20)&$Q_3$(\faCheck,0.39)&$Q_4$(\faCheck,0.55)&$Q_5$(\faCheck,0.56)&$Q_6$(\faCheck,0.65)&$Q_7$(\faTimes,0.52)&$Q_8$(\faCheck,0.61)&$Q_9$(\faCheck,0.56)&$Q_{10}$(\faCheck,0.77)\\
		\hline
		$S_2$&$Q_1$(\faTimes,0.38)&$Q_2$(\faCheck,0.50)&$Q_3$(\faTimes,0.51)&$Q_4$(\faCheck,0.54)&$Q_5$(\faCheck,0.64)&$Q_6$(\faCheck,0.65)&$Q_7$(\faTimes,0.28)&$Q_8$(\faCheck,0.61)&$Q_9$(\faTimes,0.52)&$Q_{10}$(\faTimes,0.56)\\
		\hline
		$S_3$&$Q_1$(\faTimes,0.40)&$Q_2$(\faCheck,0.23)&$Q_3$(\faCheck,0.32)&$Q_4$(\faTimes,0.29)&$Q_5$(\faTimes,0.49)&$Q_6$(\faTimes,0.47)&$Q_7$(\faTimes,0.44)&$Q_8$(\faCheck,0.52)&$Q_9$(\faCheck,0.58)&$Q_{10}$(\faTimes,0.65)\\
		\hline
		$S_4$&$Q_1$(\faTimes,0.32)&$Q_2$(\faCheck,0.44)&$Q_3$(\faTimes,0.44)&$Q_4$(\faTimes,0.43)&$Q_5$(\faTimes,0.45)&$Q_6$(\faCheck,0.45)&$Q_7$(\faTimes,0.75)&$Q_8$(\faCheck,0.78)&$Q_9$(\faCheck,0.66)&$Q_{10}$(\faCheck,0.51)\\
		\hline
		$S_5$&$Q_1$(\faCheck,0.45)&$Q_2$(\faCheck,0.65)&$Q_3$(\faCheck,0.69)&$Q_4$(\faTimes,0.79)&$Q_5$(\faTimes,0.82)&$Q_6$(\faCheck,0.86)&$Q_7$(\faTimes,0.82)&$Q_8$(\faTimes,0.75)&$Q_9$(\faTimes,0.82)&$Q_{10}$(\faTimes,0.86)\\
		\hline
		\multicolumn{11}{c}{Optimized Cognitive Representation}\\
		\hline
		$S_1$&$Q_1$(\faCheck,0.15)&$Q_2$(\faCheck,0.20)&$Q_3$(\faCheck,0.39)&$Q_4$(\faCheck,0.55)&$Q_5$(\faCheck,0.56)&$Q_6$(\faCheck,0.65)&$Q_7$(\faTimes,0.52)&$Q_8$(\faCheck,0.61)&$Q_9$(\faCheck,0.56)&$Q_{10}$(\faCheck,0.77)\\
		\hline
		$S_2$&$Q_1$(\faTimes,0.38)&$Q_2$(\faCheck,0.50)&$Q_3$(\faTimes,0.51)&$Q_4$(\faCheck,0.54)&$Q_5$(\faCheck,0.64)&$Q_6$(\faCheck,0.65)&$Q_7$(\faTimes,0.28)&$Q_8$(\faCheck,0.61)&$Q_9$(\faTimes,0.52)&$Q_{10}$(\faTimes,0.56)\\
		\hline
		$S_3$&$Q_1$(\faTimes,0.40)&$Q_2$(\faCheck,0.23)&$Q_3$(\faCheck,0.32)&$Q_4$(\faTimes,0.29)&$Q_5$(\faTimes,0.49)&$Q_6$(\faTimes,0.47)&$Q_7$(\faTimes,0.44)&$Q_8$(\faCheck,0.52)&$Q_9$(\faCheck,0.58)&$Q_{10}$(\faTimes,0.65)\\
		\hline
		$S_4$&$Q_1$(\faTimes,0.32)&$Q_2$(\faCheck,0.44)&$Q_3$(\faTimes,0.44)&$Q_4$(\faTimes,0.43)&$Q_5$(\faTimes,0.45)&$Q_6$(\faCheck,0.45)&$Q_7$(\faTimes,0.75)&$Q_8$(\faCheck,0.78)&$Q_9$(\faCheck,0.66)&$Q_{10}$(\faCheck,0.51)\\
		\hline
		$S_5$&$Q_1$(\faCheck,0.45)&$Q_2$(\faCheck,0.65)&$Q_3$(\faCheck,0.69)&$Q_4$(\faTimes,0.79)&$Q_5$(\faTimes,0.82)&$Q_6$(\faCheck,0.86)&$Q_7$(\faTimes,0.82)&$Q_8$(\faTimes,0.75)&$Q_9$(\faCheck,0.82)&$Q_{10}$(\faTimes,0.86)\\
		\hline
	\end{tabular}
\end{table*}
In terms of the final answers, if a student correctly answers a few very high difficulty questions, it means that the student seems to have mastered and become skilled at that question and has a deep understanding of it. Then, the very low-difficulty questions could probably have been written correctly, but the student's cognition was not adequately expressed at the initial stage due to distractors (slipping , questioning habits, and the structure of the question itself), resulting in a dissonance of cognitive representations. Conversely, if a student answers more difficult questions correctly in the initial stages but later repeatedly makes errors on easier questions, this suggests the presence of contingent factors, such as guessing, or non-proficiency-related factors (e.g., question structure or answering habits), which may not accurately reflect the student’s true level of competence. Therefore, it is likely that the student’s cognition is still developing but is affected by interferences at the initial stage, leading to unstable performance and ultimately resulting in a dissonance in cognitive representations. As can be seen from the students' $S_4$ practice questions, most of the exercises were at a similar level of difficulty, with only some variation in the difficulty of individual exercises. In terms of variation, for example in students' performance between questions Q20 and Q25, these questions were of similar difficulty. Before  treatment, nearly 3/5 of these questions did not perform well, and past and future questions of similar difficulty, though farther apart, did not perform well. However, the original record was not effective in expressing the lack of synergy in cognitive representations caused by the intervention of distractors. After the treatment, it can be seen that the performance of students' $S_4$ interaction records tends to be synergistic and consistent, reducing the dissonance caused by distractors. As shown in Table \ref{table20}, we selected the first 10 questions from the 30 questions for each student and compared the distributions of their performance before and after optimization. $Q(r, d)$ represents the student’s response to a question, where $r$ indicates whether the answer is correct or incorrect, and $d$ denotes the difficulty of the question. Taking student $S_1$ as an example, we can observe a significant imbalance between the student's performance at the beginning and later on from answer distribution is shown in Table \ref{table20}. The student's responses to simple questions like $Q_1$ and $Q_2$ are inconsistent with those to the more difficult questions that follow. Initially, the student answers very simple questions incorrectly but later performs exceptionally well on more challenging problems. This suggests that the early mistakes may have been caused by distractions, such as a slip, which led to incorrect answers. As a result, the model might mistakenly interpret the student's understanding of the question as inadequate or unfamiliar.

\subsection{Ablation Study }
\label{section5(d)}
In this section, we conduct ablation experiments mainly to discuss the validity and connectivity of the modules of our proposed model, as shown in Table \ref{table3}. 
\begin{itemize}
	\item {\textbf{DKT+Coo}}: Coordination module only.
	\item {\textbf{DKT+Col}}: Collaboration module only.
	\item {\textbf{DKT+Bte}}: Training embeddings with bipartite graph only.
	\item {\textbf{DKT+Bte+Coo}}: Only training embeddings with bipartite graph training, and coordination module.
	\item {\textbf{DKT+Bte+Col}}: Only training embeddings with bipartite graph training, and collaboration module.
	\item {\textbf{DKT+Coo+Col}}: Only coordination module and collaboration module.
\end{itemize}
\par

\begin{table}[H]
	\caption{ Results of the ablation study.}   
	\vspace{-0.6em}                  
	\centering
	\scalebox{0.95}{
	 	\begin{tabular}{ccccccc}                         
	 	\hline                                
	 	\multirow{2}{*}{Model}&                      
	 	\multicolumn{2}{c}{\textbf{ASSIST09}}&            
	 	\multicolumn{2}{c}{\textbf{ASSIST12}}&
	 	\multicolumn{2}{c}{\textbf{EdNet}}\cr     
	 	&AUC &ACC&AUC &ACC &AUC &ACC\\
	 	\hline                                      
	 	DKT&0.7356&  0.7179&	0.7013&  0.6842&	0.6909&  0.6889\\
	 	\hline 
	 	DKT+Coo&0.7869&  0.7526&	0.7388&  0.7403&	0.7133&  0.6706\\
	 	DKT+Col&0.7721&  0.7375&	0.7423&  0.7289&	0.7132&  0.6815\\
	 	DKT+Bte&0.8287&  0.7669&	0.7665&  0.7423&	0.7765&  0.7076\\
	 	\hline 
	 	DKT+Bte+Coo&0.8567&  0.8024&	0.7807&  0.7651&	0.7806&  0.7140\\
	 	DKT+Bte+Col&0.8503&  0.7870&	0.7802&  0.7583&	0.7810&  0.7189\\
	 	DKT+Coo+Col&0.8042&  0.7720&	0.7690&  0.7561&	0.7213&  0.6860\\
	 	\hline 
	 	CRO-KT&0.8782&  0.8183&	0.8051&  0.7812&	0.7880&  0.7258\\
	 	\hline                                    
	 	\label{table3}
	 \end{tabular}
}                                     

\end{table}

\begin{table*}[h]
	\caption{Quantified Optimization Results of Three Datasets}
	\centering     
	\label{tab:freq}
	\begin{tabular}{cccccccccccccccl}
		\hline
		ASSIST09(Coo)&AUC&ACC&ASSIST12(Coo)&AUC&ACC&EdNet(Coo)&AUC&ACC\\
		\hline
		0$\%$ & 0.8287& 0.7669&0$\%$ &0.7665& 0.7423&0$\%$&0.7765& 0.7076\\
		30$\%$& 0.8485&	0.7854&30$\%$ & 0.7765& 0.7723&30$\%$&0.7799&0.7126\\
		50$\%$& 0.8546&	0.7966&50$\%$ & 0.7800& 0.7788&50$\%$&0.7808& 0.7188\\
		70$\%$& 0.8574&	0.8006&70$\%$ & 0.7854& 0.7791&70$\%$&0.7862& 0.7194\\
		\hline
		ASSIST09(Col)&AUC&ACC&ASSIST12(Col)&AUC&ACC&EdNet(Col)&AUC&ACC\\
		\hline
		0$\%$ & 0.8287& 0.7669&0$\%$ &0.7665& 0.7423&0$\%$&0.7765& 0.7076\\
		30$\%$& 0.8456&	0.7844&30$\%$ &0.7754& 0.7711&30$\%$&0.7800&0.7123\\
		50$\%$& 0.8486&	0.7849&50$\%$ &0.7789& 0.7799&50$\%$&0.7818& 0.7180\\
		70$\%$& 0.8540&	0.8001&70$\%$ & 0.7801& 0.7759&70$\%$&0.7828& 0.7200\\
		\hline
		ASSIST09(Coo+Col)&AUC&ACC&ASSIST12(Coo+Col)&AUC&ACC&EdNet(Coo+Col)&AUC&ACC\\
		\hline
		0$\%$ & 0.8287& 0.7669&0$\%$ &0.7665& 0.7423&0$\%$&0.7765& 0.7076\\
		30$\%$& 0.8521&	0.7888&30$\%$ &0.7832&0.7766&30$\%$&0.7811&0.7204\\
		50$\%$& 0.8601&	0.8034&50$\%$ &0.7901& 0.7822&50$\%$&0.7866& 0.7222\\
		70$\%$& 0.8680&	0.8013&70$\%$ & 0.7934& 0.7900&70$\%$&0.7870& 0.7244\\
		\hline
		\label{table4}
	\end{tabular}
\end{table*}

\textemdash From the original CRO-KT module we strip other original modules and let the remaining modules show their effectiveness. Compared with DKT model, DKT+Coo and DKT+Col have significant contribution, which shows that the original records have great limitation on the expression of students' cognition. The optimization of the original records by our model effectively reduces the incoherence and lack of synergy in the records, which shows the effectiveness of the optimization.\par
\textemdash In contrast to DKT+Bte, which only incorporates the bipartite graph training embedding, DKT+Bte+Coo and DKT+Bte+Col have a larger contribution. This suggests that the approach of constructing relationships between different problems is still limited by the original record, and also demonstrates the feasibility and effectiveness of optimizing the record.\par
\textemdash From the experimental results, DKT+Coo+Col has a more significant contribution compared to the results of DKT+Coo. This suggests suggests that it is not sufficient to consider only the case of incoherence between exercises; it is also important to address the issue of lack of synergy. This verifies the validity and correctness of considering both cases simultaneously.\par

\textemdash Taking into account the implicit relationships between different knowledge points, as well as the effective representation of the various questions answered by students, helps improve the accuracy of the model. Based on this implicit relationship, there exists an optimal combination of representations between exercises. The significant contribution of CRO-KT can be seen from DKT+Bte and CRO-KT, where the CRO-KT model is equivalent to DKT+Bte+ Coo+Col, which further validates the effectiveness of our optimization algorithm to significantly represent cognitive representations.\par

\begin{figure*}[h]
	\centering
	\includegraphics[width=17cm,height=6.6cm]{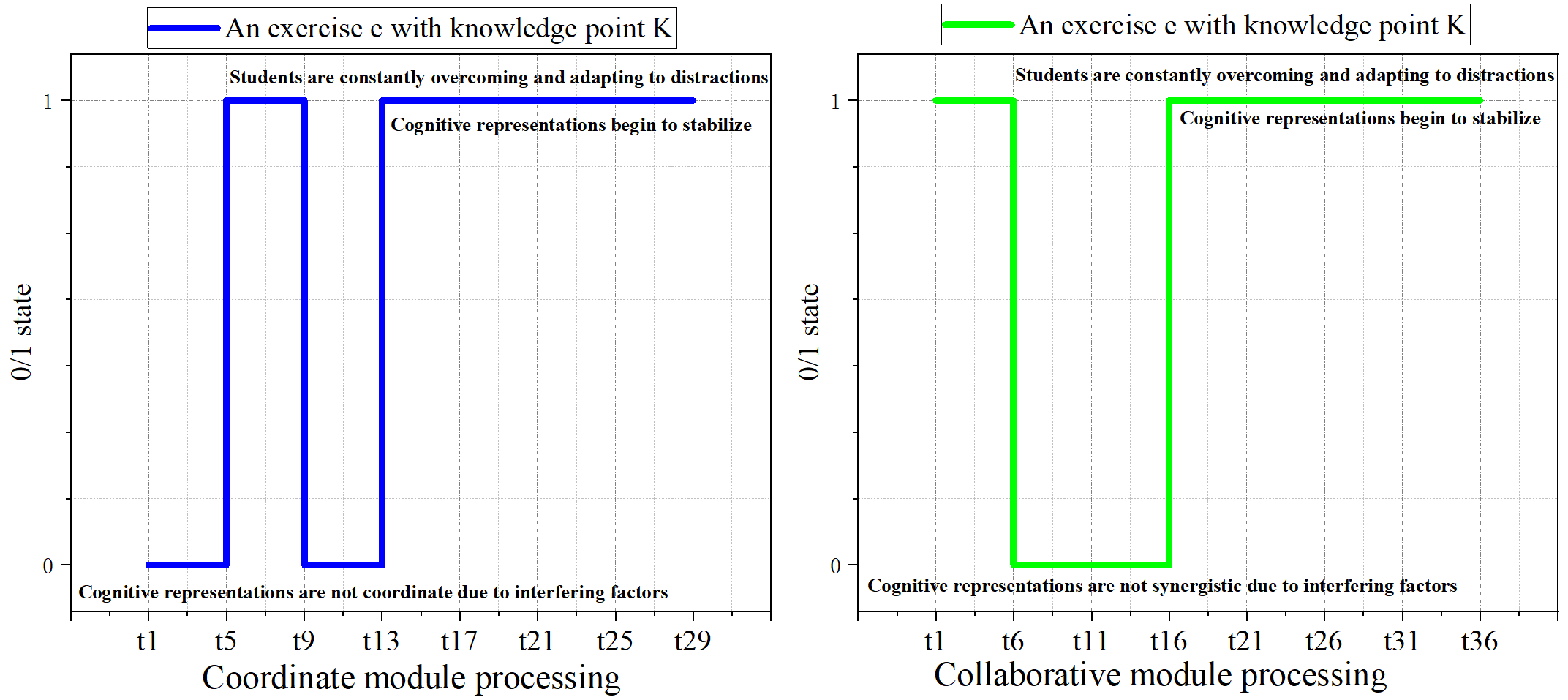}
	\caption{ Performance of exercise $e$ in modules.}
	\label{fig10}
\end{figure*}

\begin{figure*}[h]
	\centering
	\includegraphics[width=17cm,height=6.6cm]{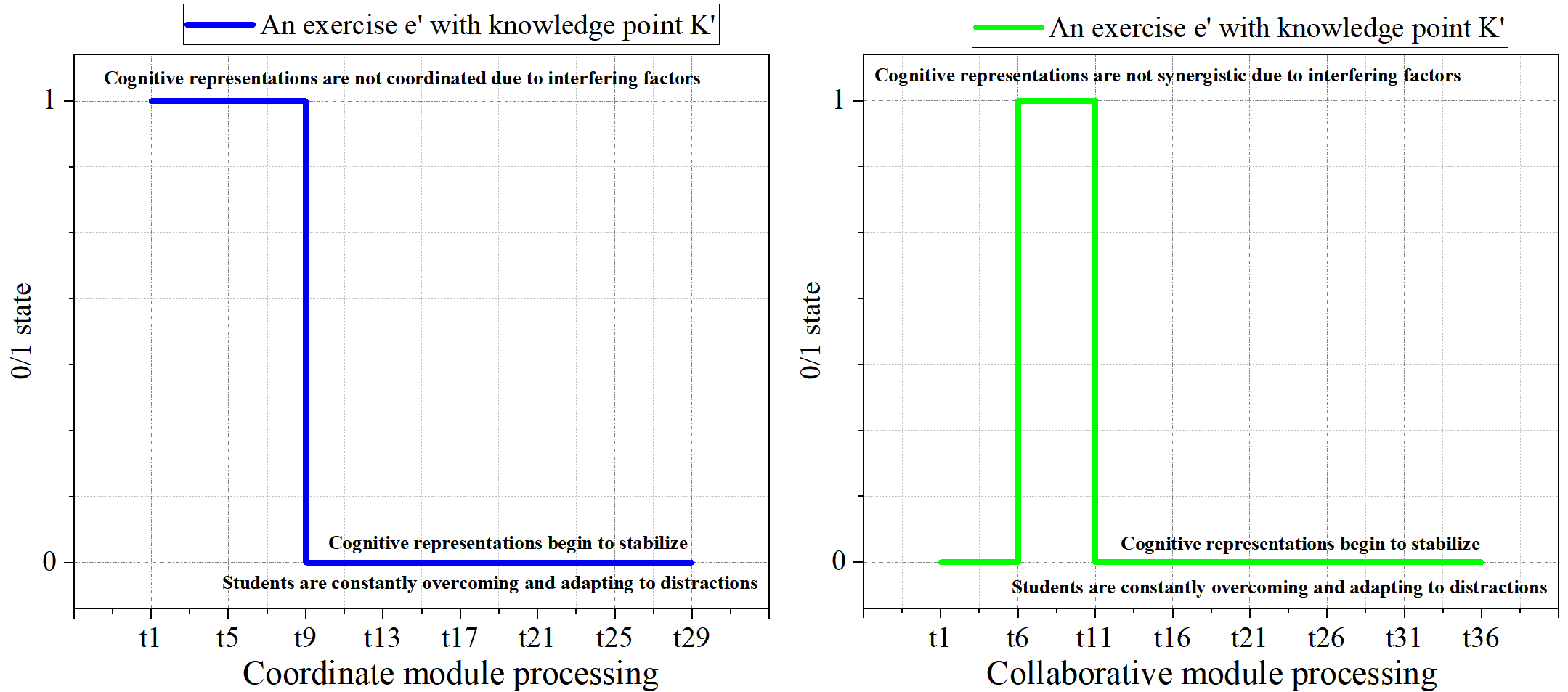}
	\caption{Performance of exercise $e'$ in modules.}
	\label{fig11}
\end{figure*}

As we have stated in our opinion, there are indeed many confounding factors in a student's performance. We feel that these may include both subjective factors (carelessness, guessing) and objective factors (the structure of the problem itself and the student's habit of doing the problem). At the same time, we believe that as students continue to adapt to and overcome these confounding factors, subsequent cognitive representations will stabilize. In other words, further cognitive representations are more representative of students' perceptions. In this regard, we conducted experiments on two exercises $e$ and $e'$, which we selected as different points of knowledge, and compared them. The results of the experiments show that in the initial phase students are susceptible to distractors. These influences lead to inauthenticity of the recorded representations (inauthenticity of correct answers and inauthenticity of incorrect answers). As the interaction increased and students adapted to and overcame the distractors, the authenticity and stability of the recording representations began to emerge. The results are shown in Fig. \ref{fig10} and Fig. \ref{fig11}. 
\begin{figure*}[h]
	\centering
	\includegraphics[width=17cm,height=13cm]{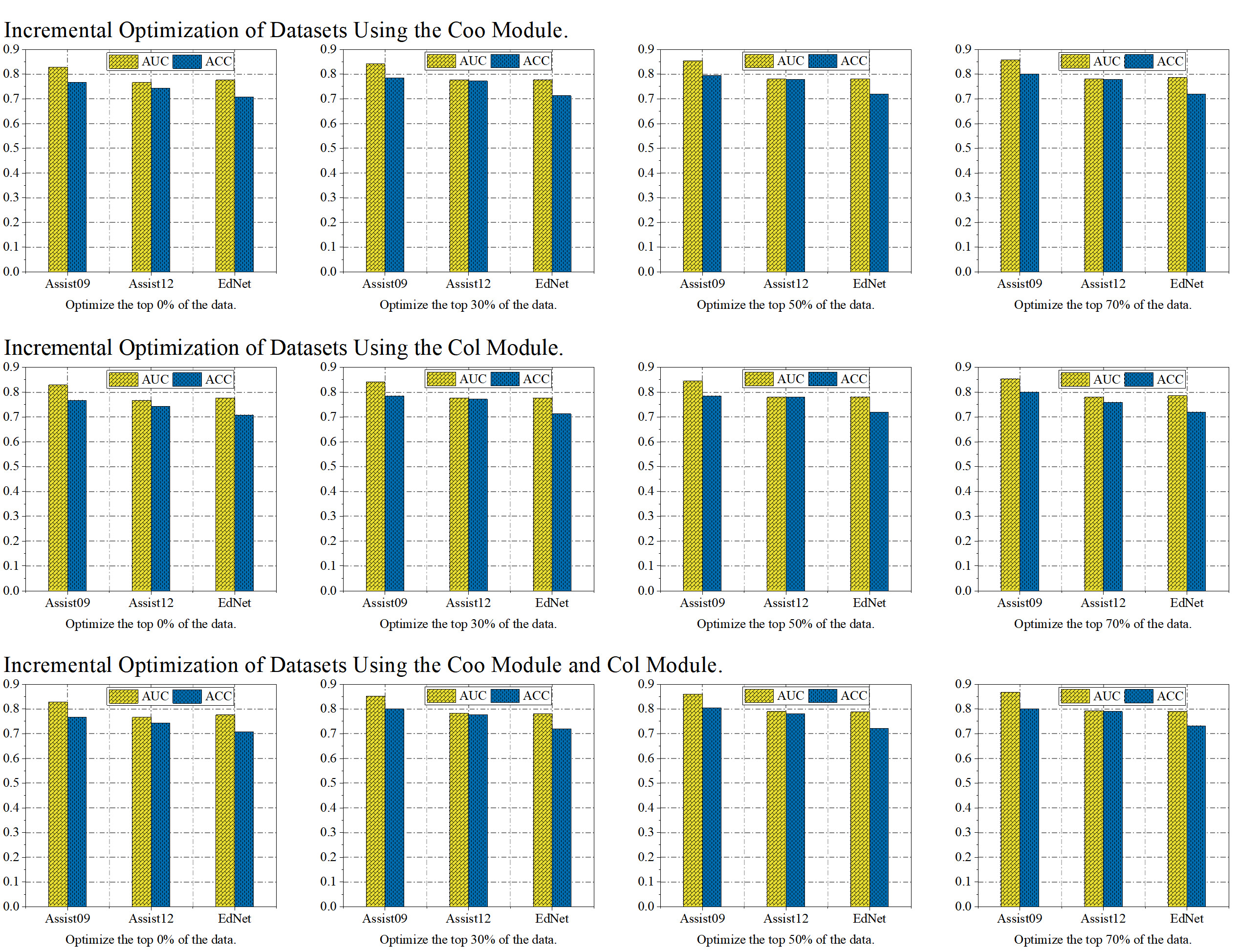}
	\caption{Data quantification experiments of different modules.}
	\label{fig12}
\end{figure*}
\subsection{Data Quantification Experiment}

\begin{figure*}[h]
	\centering
	\includegraphics[width=17cm,height=4.6cm]{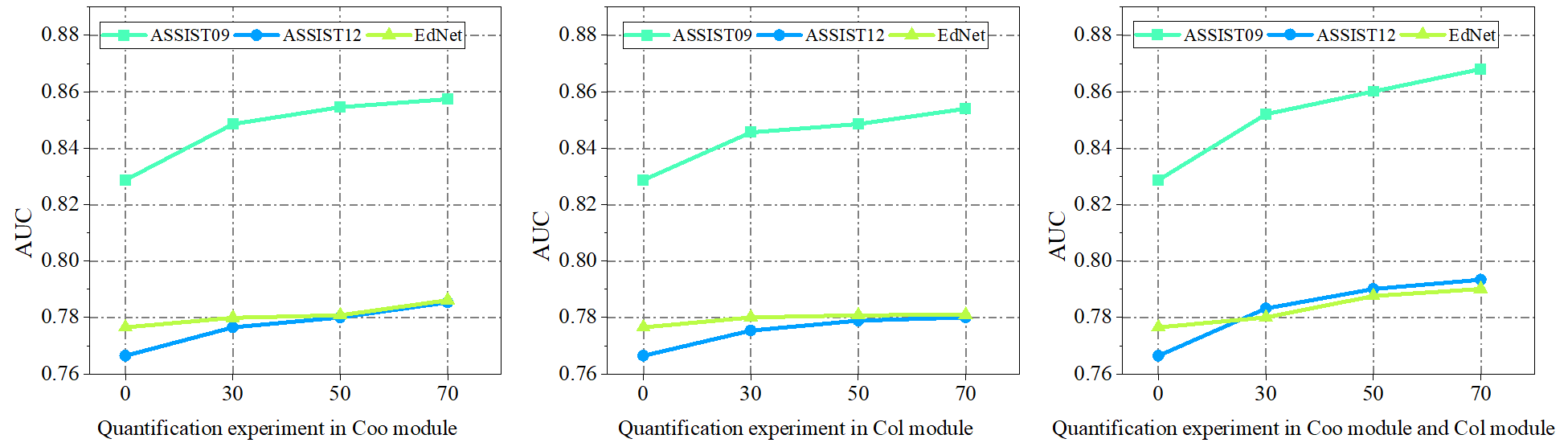}
	\caption{Line chart of performance improvement for different modules with varying proportions of optimized data.}
	\label{fig13}
\end{figure*}
The CRO-KT model's task primarily involves identifying the best representations that align closely with students' cognition. To verify that our optimized data is highly correlated with subsequent responses, we conducted experiments by extracting the top 30$\%$, 50$\%$, and 70$\%$ of the data from the sequence, comparing it with unoptimized data, where the performance of the unoptimized data serves as a benchmark represented by DKT+Bte(see Section \ref{section5(d)}). As shown in Fig. \ref{fig12}, the experiments are divided into three groups. The first row presents the quantification experiments conducted solely by the coordination module, the second row depicts the quantification experiments of the collaboration module, and the third row displays the quantification experiments with both the coordination and collaboration modules combined. The experimental results indicate that performance tends to increase gradually as the amount of optimized data expands. This suggests that even a small amount of optimization can maintain a high correlation with subsequent response situations, without exhibiting significant information inconsistencies or contradictions, while also ensuring data quality. As illustrated in Fig. \ref{fig13}, it is evident that the growth rates of optimization at different proportions vary, generally decreasing from high to low before gradually stabilizing. This suggests that there are many optimization targets needed for students' initial sequences, reflecting the fact that students in the early stages of learning are easily distracted, leading to insufficient expression. As students gradually adapt, their cognitive state stabilizes, causing performance improvements to slow down. Our data quantification experimental results are presented in Table \ref{table4}.

\subsection{Parameter Sensitivity Experiments}
According to heuristic rules, we set the parameters $\alpha$ and $\beta$ within a specific range. Then, using random search, we randomly select parameter values within this range in an increasing order. For each parameter value, we calculate the difference in performance between it and its neighboring parameter values. If the difference is smaller, it indicates higher robustness. We choose the parameter with the smallest difference and the highest robustness from these values.\par
\begin{figure*}[h]
	\centering
	\includegraphics[width=17cm,height=6.6cm]{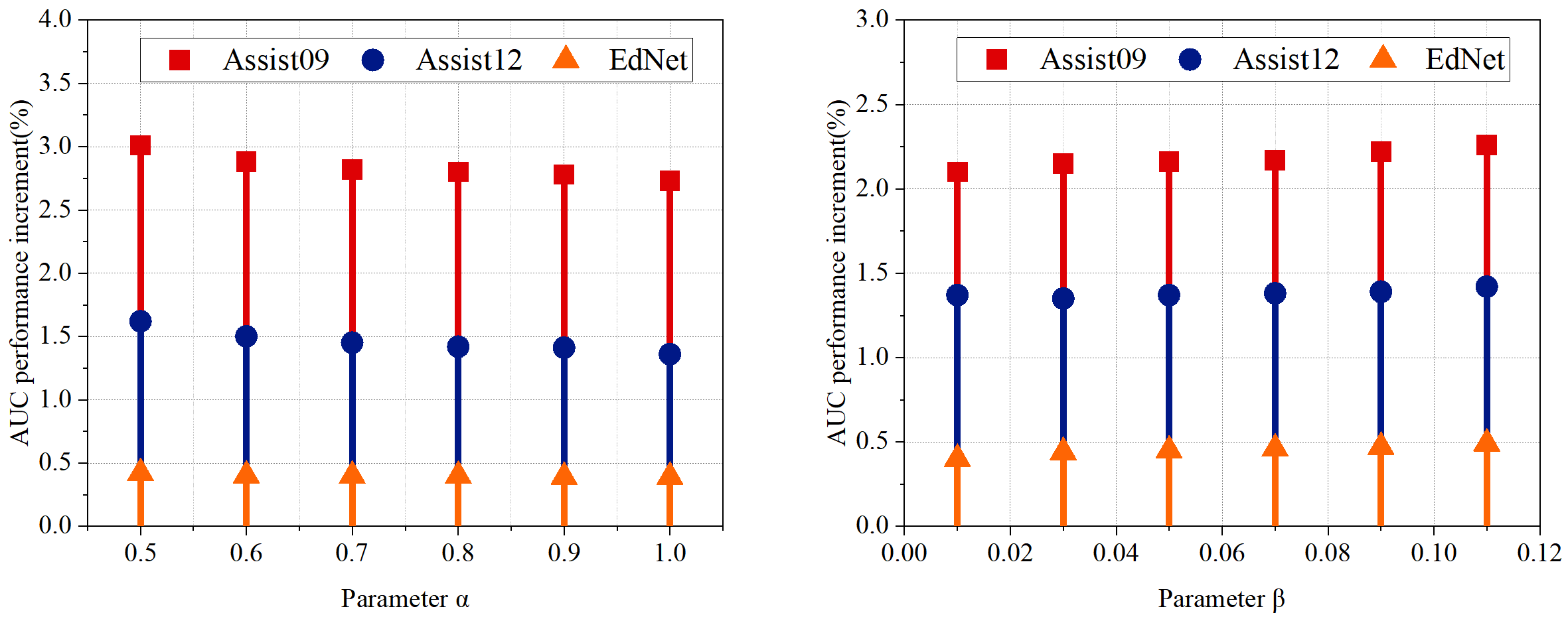}
	\caption{Sensitivity experiments of parameter $\alpha$  and  $\beta$.}
	\label{fig14}
\end{figure*}

The two key parameters $\alpha$ and $\beta$ set by the CRO-KT model are based on the coordination and collaboration modules' adjustable parameters, respectively. In order to show the changes in the performance of each module in our parameter tuning, we conduct parameter sensitivity experiments based on the performance of DKT+Bte for comparison (see Section \ref{section5(d)}). Based on the parameter $\alpha$ of the coordination module, it represents the absolute value of the difference in difficulty between two exercises, where the knowledge points of the two exercises are the same. Theoretically, the larger  difference in difficulty, the more reasonable it is, so we choose to adjust the parameter $\alpha$ in the range where the difference in difficulty is larger, as shown in the left panel of Fig. \ref{fig14}, where the parameter $\alpha$ is chosen in the range of 0.5-1.0. The horizontal coordinate represents the difference in difficulty between the two problems, and the vertical coordinate represents the incremental AUC performance of the model that can be achieved with the corresponding parameter $\alpha$. Our method is intended to find out whether there is coordination between the two exercises. The absolute value of the difference in difficulty between the two exercises is greater than or equal to this parameter $\alpha$, so the larger and more overall robust parameter $\alpha$ is the desired parameter metric. As can be seen from the figure, the parameter $\alpha$ is set around 0.8, which shows a certain level of robustness in performance.

Based on the parameter $\beta$ of the collaboration module, which represents the absolute value of the difference in difficulty between the two exercises that share the same knowledge points. Theoretically, a smaller difficulty difference is more reasonable, so we choose to adjust the parameter $\beta$ in the range where the difficulty difference of the exercises is smaller. It is important to note that the similarity in difficulty between problems inherently involves a certain degree of error, and an optimal range of error can lead to more favorable outcomes. In practical scenarios, the equality of problem difficulty essentially refers to a minimal difference in the level of difficulty between the problems. As shown in the right panel of Fig. \ref{fig14}, the parameter $\beta$ is selected in the range of 0.01-0.11. The horizontal coordinate represents the difficulty difference $\beta$, and the vertical coordinate represents the incremental AUC performance of the model that can be achieved with the corresponding parameter $\beta$. The collaboration module aims to find topics with the same knowledge points and very similar difficulty, where the absolute value of the difficulty difference between the two exercises is less than or equal to this parameter $\beta$. A smaller $\beta$ generally results in better performance and greater robustness, making it the parameter index we need. From the experimental results, the parameter $\beta$ is set at about 0.05 and the performance shows a certain robustness.\par

\section{Conclusion}

In this paper, we propose a model called CRO-KT to optimize students' cognitive representations. Unlike traditional models, this model focuses more on the accurate representation of raw records and reduces ineffective correlations and disorganized structures within the records through optimization algorithms. It also constructs a bipartite graph to generate relational embeddings, which are then combined with the optimized records to further enhance cognitive representations. We validated the model's performance on three publicly available datasets and compared it with seven excellent methods. Experimental results show that the proposed method achieves state-of-the-art performance. However, despite the model's outstanding performance in experiments, its application in real-world educational systems still faces certain challenges. Future research could explore how to extend the CRO-KT model to more diverse educational tasks, such as personalized learning path recommendations, student behavior prediction, and cross-disciplinary knowledge integration. Additionally, incorporating non-cognitive factors (such as students' emotional states, learning motivations, etc.) into cognitive representations may further improve the model's practical effectiveness and generalizability. For future work, we plan to further optimize the model's computational efficiency, particularly for applications in large-scale educational datasets, and investigate students' non-cognitive factors by integrating interval performance information or incorporating these non-cognitive factors into graph embeddings as hidden or overlayed information.

\bibliographystyle{IEEEtran}
\bibliography{reference}     
\vspace{-12mm}

\vfill

\end{document}